\documentclass{article}

\PassOptionsToPackage{numbers, compress, sort}{natbib}

    \usepackage[preprint]{neurips_2025}

\usepackage[utf8]{inputenc} 
\usepackage[T1]{fontenc}    
\usepackage{hyperref}       
\usepackage{url}            
\usepackage{booktabs}       
\usepackage{amsfonts}       
\usepackage{nicefrac}       
\usepackage{microtype}      
\usepackage[table]{xcolor}  
\usepackage[sort,compress]{natbib} 

\usepackage{graphicx}
\graphicspath{ {./figures/} }

\usepackage{multirow}
\usepackage{array}
\usepackage{amsmath}
\usepackage{enumitem}

\usepackage[compact]{titlesec}
\titlespacing{\section}{0pt}{*0.3}{*0}
\titlespacing{\subsection}{0pt}{*0.15}{*0}

\usepackage[subtle, mathdisplays=tight, charwidths=normal, leading=normal]{savetrees}

\usepackage{listings}
\usepackage{placeins}  
\definecolor{lightgray}{RGB}{245,245,245}
\definecolor{pycomment}{RGB}{224,0,136}  
\definecolor{pykeyword}{rgb}{0.1,0.1,0.6}
\lstdefinestyle{algo}{
    language=Python,                 
    backgroundcolor=\color{lightgray},
    basicstyle=\ttfamily\small,
    commentstyle=\color{pycomment},
    keywordstyle=\color{pykeyword}\bfseries,
    showstringspaces=false,
    frame=single,                    
    rulecolor=\color{black!40},
    frameround=tttt,
    xleftmargin=0pt,
    framexleftmargin=0pt,
    framexrightmargin=0pt,
    framextopmargin=6pt,
    framexbottommargin=6pt,
    aboveskip=1em,
    belowskip=1em,
}

\title{Chipmunk: Training-Free Acceleration of Diffusion Transformers with Dynamic Column-Sparse Deltas}

\author{%
  Austin Silveria\\
  University of California, San Diego\\
  Together AI\\
  \texttt{austin@together.ai} \\
   \And
  Soham V. Govande \\
  Stanford University \\
  \texttt{govande@stanford.edu} \\
     \And
    Daniel Y. Fu \\
  University of California, San Diego \\
  Together AI\\
  \texttt{danfu@ucsd.edu} \\
}

\begin{document}
\graphicspath{ {./figures/} }

\maketitle

\begin{abstract}
Diffusion Transformers (DiTs) have achieved state-of-the-art performance in high-quality image and video generation but incur substantial compute cost at inference.
A common observation is that DiT latent noise vectors change slowly across inference steps, which suggests that the DiT compute may be redundant across steps.
In this paper, we aim to speed up inference by reducing this redundancy, without additional training. 
We first study how activations change between steps in two state-of-the-art open-source DiTs.
We find that just 5-25\% of the values in attention and MLP explain 70-90\% of the change in activations across steps.
This finding motivates our approach, Chipmunk, which uses dynamic sparsity at inference time
to recompute only the fastest-changing intermediate activations, while caching the rest.
Dynamic sparsity introduces two systems challenges:
(1) sparse attention and MLP operations tend to underutilize GPU tensor cores;
and (2) computing dynamic sparsity patterns at runtime and caching activations both introduce overhead.
To address these challenges, Chipmunk first uses a voxel-based reordering of input tokens to introduce \textit{column-wise} sparsity.
We implement column-sparse kernels utilizing efficient sparse gathers from global to shared GPU memory, achieving a 9.3x speedup at 93\% sparsity compared to highly-optimized dense baselines.
Second, Chipmunk overlaps the computation of sparsity patterns and cache updates with other parts of the computation (e.g., second layer of the MLP) to hide the extra latency. 
Chipmunk achieves up to 2.16x speedup on HunyuanVideo and 1.41x on FLUX.1-dev without compromising generation quality.
Furthermore, we show that Chipmunk can be stacked on top of full step caching, achieving a 3.72x speedup on HunyuanVideo, a 2.67x speedup on WAN2.1, and a 2.25x speedup on FLUX.1-dev with minimal quality impact. 
 
\end{abstract}
\section{Introduction}

Diffusion Transformers (DiTs) have emerged as state-of-the-art (SOTA) models for generating high-quality images and videos \cite{peebles2023scalablediffusionmodelstransformers,zheng2024opensorademocratizingefficientvideo,hong2022cogvideolargescalepretrainingtexttovideo,kong2024hunyuanvideo,flux2024,chen2023pixartalphafasttrainingdiffusion}. However, DiTs are increasingly constrained by their substantial computational requirements as sequence lengths and parameters scale up \cite{sun2024unveilingredundancydiffusiontransformers,yao2024fasterditfasterdiffusiontransformers}. For example, HunyuanVideo, with a 118k sequence length and 13B parameters, requires 18 minutes to generate a 5s video on an H100 GPU \cite{kong2024hunyuanvideo}. In particular, two major redundancies contribute significantly to unnecessary computation: the slow-changing latent vector (DiT model input) \cite{ma2024deepcache,sun2024unveilingredundancydiffusiontransformers,yuan2024ditfastattnattentioncompressiondiffusion} across generation steps, and the inherent sparsity in DiT activations \cite{li2022lazy,liu2024training}.

Existing approaches leverage these redundancies by caching per-step \cite{liu2024timestep}, per-layer \cite{wimbauer2024cachecanacceleratingdiffusion}, or per-token \cite{zou2025acceleratingdiffusiontransformerstokenwise} outputs, or by computing sparse attention with static local window techniques \cite{zhang2025fastvideogenerationsliding,yuan2024ditfastattnattentioncompressiondiffusion}. In this paper, we ask whether we can achieve greater training-free acceleration by exploiting this redundancy \textit{dynamically}, in the \textit{most granular} manner possible.

We begin with an initial quantitative analysis of activation patterns in two state-of-the-art open source DiTs (HunyuanVideo and FLUX.1-dev). We find that over 90\% of the variance in cross-step attention activation changes is explained by only 5-25\% of the intermediate activation values (Table \ref{tab:sparsity_comparison}). That is, recomputing only the top 5-25\% of attention interactions and reusing the rest from the previous step captures over 90\% of the cross-step change. For MLPs, we find 15-25\% of the intermediate activation values explain over 70\% of the variance (Table \ref{tab:sparsity_comparison}).

\begin{figure}
    \centering
    \includegraphics[width=1\linewidth]{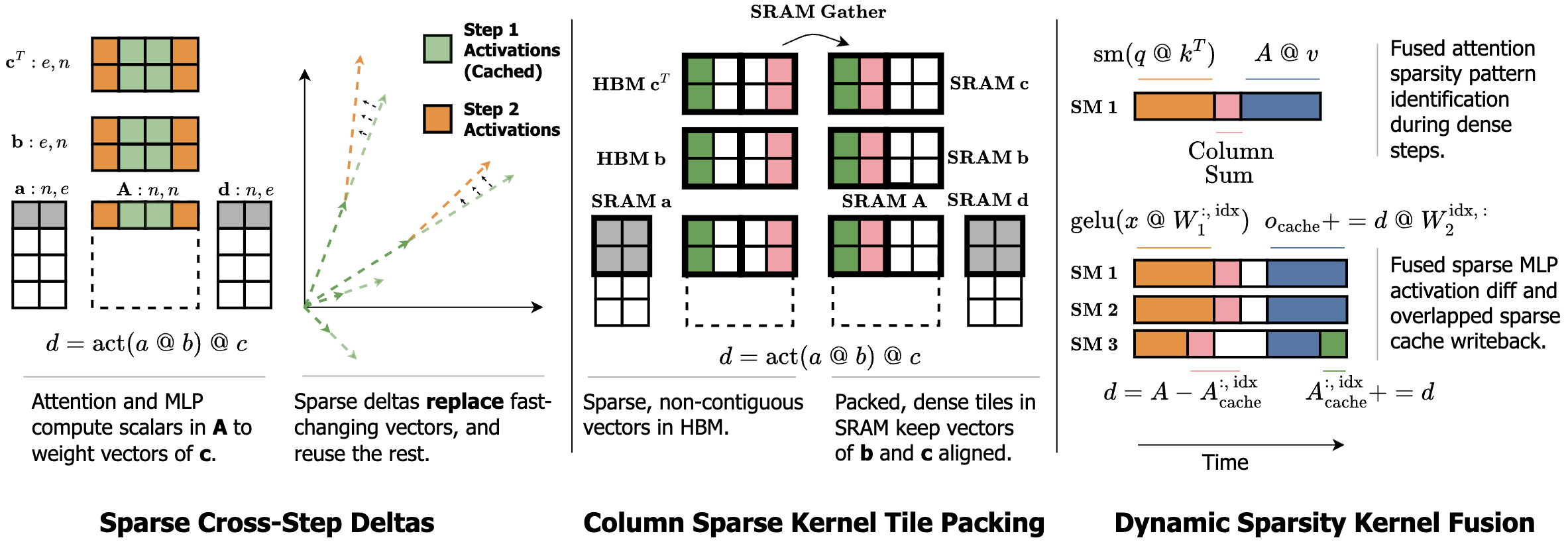}
    \caption{\textit{Left:} Chipmunk uses intermediate activation sparsity to recompute only the fastest changing vectors in the output linear combinations of attention and MLPs. \textit{Middle:} Column sparse kernels achieve low approximation error and hardware-efficiency by packing dense SRAM tiles for peak tensor core utilization. \textit{Right:} Extra operations required to compute dynamic sparse deltas, such as sparsity pattern identification and cache updates, are fused with other operations to minimize overhead.}
    \label{fig:delta-packing-fusion}
\end{figure}

\textbf{Chipmunk.}
This finding motivates our method, Chipmunk, which caches intermediate activations and uses dynamic sparsity to recompute only the fastest-changing activations.
We exploit a common computational structure of attention and MLPs--$\text{act}(a \; @ \; b) \; @ \; c$--both using a back-to-back matrix multiply to produce a linear combination of vectors in $c$ (Section~\ref{background}). Sparsity on the intermediate activations of both operations corresponds one-to-one with these individual vector contributions.
Specifically, Chipmunk uses the magnitudes of intermediate activations to determine which vectors of $c$ to recompute at each step, and uses cached values of the rest of the vectors from the previous step.
For attention, Chipmunk chooses the top-k \textit{column-chunks} of the attention matrix to recompute, and for MLP, Chipmunk computes an approximate difference against cached activations to identify the top-k column-chunks to recompute. 
This can be interpreted intuitively as computing sparse deviations from straight line paths in latent space (Fig. \ref{fig:delta-packing-fusion}, left), where the recomputed vectors are dynamically allocated more sequential computation steps.


\textbf{Systems Challenges.}
Dynamic sparsity introduces two systems challenges:
(1) Sparse attention and MLP operations tend to underutilize tensor cores, which can make it challenging to achieve wall-clock speedup relative to dense baselines.
(2) Computing dynamic sparsity patterns at runtime and caching activations introduces additional overhead.

To address these challenges, Chipmunk uses a voxel-based ordering of the pixels (Fig. \ref{fig:attn-mlp-error}, left) to regroup spatiotemporally local tokens into \textit{column-wise} sparsity patterns (Fig. \ref{fig:delta-packing-fusion}, middle).
In particular, a contiguous group of tokens activates the \textit{same} sparse set of individual keys/values (attention) or weights (MLPs). This corresponds to selecting columns of $K^T$/$W_1$ and rows of $V$/$W_2$.
This column-wise sparsity pattern admits efficient attention and MLP kernels
using sparse gathers from GPU global memory (HBM) to packed dense tiles in GPU shared memory (SRAM), which can then fully utilize GPU tensor cores (Fig. \ref{fig:delta-packing-fusion}, middle) \cite{nvidiagpuarchitecture,chen2021colsparse,Li_2022colsparse,ye2025flashinfer}.
To reduce the overhead of computing dynamic sparsity patterns and maintaining an activation cache, Chipmunk
uses custom kernels to overlap sparsity pattern identification and cache I/O operations with other computations (Fig. \ref{fig:delta-packing-fusion}, right).
Chipmunk's column-sparse attention and MLP kernels scale linearly with sparsity with little overhead, achieving nearly-optimal speedup while providing 2x less approximation error than block sparsity. At 93\% sparsity, our column-sparse attention kernel is 9.3x faster than FlashAttention-3 in the ThunderKittens library \cite{shah2024flashattention3fastaccurateattention,spector2024thunderkittenssimplefastadorable}.

\textbf{Evaluation.}
We evaluate Chipmunk on state-of-the-art text-to-video generation models (HunyuanVideo and WAN2.1) and one state-of-the-art text-to-image generation model (FLUX.1-dev).
\begin{itemize}[leftmargin=*,nosep,nolistsep]

\item \textbf{Quality.} Chipmunk attains up to 92\% attention sparsity across 44 of 50 generation steps on HunyuanVideo with minimal impact to VBench scores. On FLUX.1-dev, Chipmunk achieves 84\% attention sparsity and 70\% MLP sparsity for 44 of 50 steps without affecting ImageReward and CLIP scores.

\item \textbf{Fast Generation.} Chipmunk alone achieves end-to-end generation speedups of 2.16x on HunyuanVideo and 1.41x on FLUX.1-dev.

\item \textbf{Stacked Acceleration.} Chipmunk natively stacks with existing caching strategies, such as step caching, leading to further acceleration—achieving speedups of 3.72x on HunyuanVideo, 2.67x on WAN2.1, and 2.25x on FLUX.1-dev.

\end{itemize}

\section{Background}
\label{background}

In this section, we discuss relevant background material on Diffusion Transformers and GPU architecture. Two key takeaways are that (1) multi-step DiT inference computes paths in latent space from noise to output and (2) GPU kernel efficiency critically depends on saturating tensor cores with large block matrix multiplications.

\subsection{Diffusion Transformers (DiTs)}
We briefly review diffusion models as learning to reverse a forward noising process, DiTs as parameterizing this reverse denoising process with transformers \cite{vaswani2023attentionneed}, and the relevant shared computational form of attention and MLPs in transformers.

\paragraph{Diffusion models \cite{sohldickstein2015deepunsupervisedlearningusing} reconstruct data from noise.} Formally, a forward diffusion process defines a Markov chain $q(z_t\mid z_{t-1})$ that gradually perturbs data $z_0$ into noise $z_T$.
A neural network—denoted $\epsilon_{\theta}(z_t,t)$—is trained to approximate the reverse denoising process, learning to generate data with an inverse mapping that iteratively reconstructs $z_0$ from $z_T$. 

\paragraph{DiTs compute per-token paths in latent space.} DiTs \cite{peebles2023scalablediffusionmodelstransformers} extend Vision Transformers \cite{dosovitskiy2021imageworth16x16words} to diffusion image and video generation by applying timestep and prompt-conditioned modulation layers to attention and multilayer perceptron \footnote{In this paper, we use the term "MLP" to refer to the two-layer perceptrons used in every major DiT architecture even though technically, MLPs can have more than two layers. We also assume an activation function of GeLU, but this can be replaced with any arbitrary function. Finally, for simplicity in formulas in the paper, we omit bias terms, but our kernels \textit{do} have full support for them.} (MLP) outputs. The concrete representation denoised by DiTs is the same as language model transformers: A set of tokens, each projected from a patch of pixels to a high-dimensional vector representation (the latent vector) \cite{peebles2023scalablediffusionmodelstransformers}. In each denoising step, the DiT takes this representation as input and computes a residual.

\paragraph{Multi-step inference paths change direction at each step.} In a single-step inference, the DiT computes one forward pass, adding its output to the latent vector before decoding into pixel space—a straight-line path from initial noise to output. Multiple steps increase expressivity by allowing each computation to depend on previous outputs, causing latent paths to change direction (Fig. \ref{fig:delta-packing-fusion}, left). However, this sequential dependency substantially increases computational costs due to the multiple forward passes of the model. Prior work has found that even with models trained for single-step denoising, generations can achieve better quality with multiple inference steps \cite{song2023consistencymodels,esser2024scalingrectifiedflowtransformers}.

\paragraph{Attention and MLPs share a computational form of $\text{act}(a \; @ \;b) \; @ \; c\,$.} Recalling the equations for attention, $\text{softmax}(q \; @ \; k^T) \; @ \; v$, and MLP, $\text{gelu}(x \; @ \; W_1) \; @ \; W_2$, we see that both operations use a non-linearity to compute the scalar coefficients for a linear combination of vectors. Each scalar in the intermediate matrix scales one output vector. In attention, the vectors are dynamic ($v$ is projected from the token representation), and in MLP, the vectors are static (rows of the $W_2$).

\subsection{GPU Architecture}

Here, we cover the most relevant aspects of GPU architecture for attention and MLPs, the "workhorse" operations of DiTs. Modern GPUs parallelize these computations by dividing the work among a number of  independent Streaming Multiprocessors (e.g., 132 SMs on an H100), each equipped with compute units and high-speed local memory (SRAM) \cite{nvidiagpuarchitecture, spector2024thunderkittenssimplefastadorable}. At a high level, attention and MLP kernels both compute two tiled matrix multiplications (GEMMs). To better understand these operations, we will describe dataflow for a tiled GEMM kernel $C=A\; @\; B$ on a modern GPU \cite{tillet2019triton}: (i) load chunks of contiguous rows in $A$ and columns in $B$ from global High-Bandwidth Memory (HBM) to SRAM, (ii) compute matrix multiplies by feeding these blocks into tensor cores, and (iii) write the resulting output block of $C$ to HBM. Both FlashAttention and high-performance GEMMs (e.g. cuBLAS) partition work into such tiles, fusing activation functions such as GELU and softmax \cite{dao2022flashattentionfastmemoryefficientexact,spector2024thunderkittenssimplefastadorable}. 
Notably, FlashAttention also fuses the second GEMM into the same kernel to avoid I/O costs \cite{dao2022flashattentionfastmemoryefficientexact}.

\paragraph{Maximizing tensor core utilization is crucial to achieving peak computational performance. }In modern GPUs, the tensor core can launch block matrix multiplies on tiles in SRAM as large as 64x256, and microbenchmarks show that at least 64x64 is needed to achieve peak throughput \cite{luo2024benchmarkingdissectingnvidiahopper}. Thus, to keep a kernel competitive, it is critical to keep tensor cores well fed with a constant pipeline of data resident in SRAM \cite{spector2024thunderkittenssimplefastadorable}.
\section{Method}
\label{method}

\begin{figure}
    \centering
    \includegraphics[width=1\linewidth]{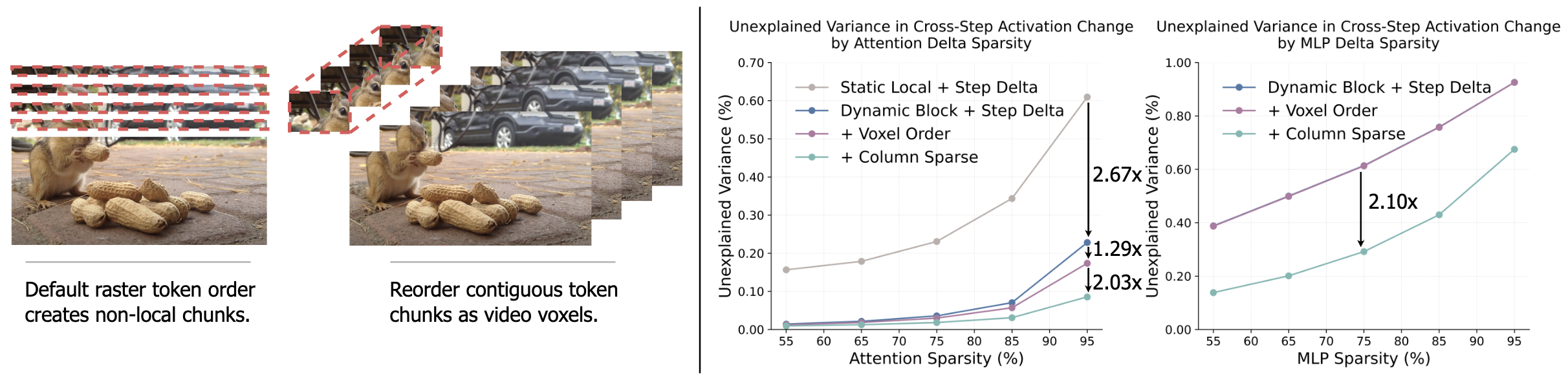}
    \caption{\textit{Left:} Chipmunk reorders tokens such that column sparse patterns route voxels (3D cubes of pixels) to the same set of activated keys/values (attention) or model weights (columns/rows of $W_1$/$W_2$ in MLPs). \textit{Right:} We plot the unexplained variance in cross-step activation changes (1 - $R^2$). Relative to dynamic block sparsity, dynamic column sparsity gives a 2x reduction in unexplained variance across both attention and MLPs. Voxel order reduces unexplained variance in attention but has nearly zero impact in MLP (overlapping with the blue line).}
    \label{fig:attn-mlp-error}
\end{figure}

To investigate the interaction between slow-changing latent noise vectors \cite{ma2024deepcache} and naturally sparse activations \cite{li2022lazy, liu2024training}, we first quantify the amount of cross-step change in attention and MLP outputs that can be captured by a sparse recomputation. This analysis motivates Chipmunk’s unified strategy of caching and fine-grained sparsity across attention and MLP layers. We then demonstrate that dynamic, fine-grained column sparsity is both hardware-efficient and achieves low approximation error. Finally, we describe Chipmunk’s end-to-end algorithm, which interleaves dense and sparse computation steps and natively stacks with coarser caching methods (e.g., per-token, per-step).

\subsection{Motivation: Sparse Cross-Step Changes in Attention and MLPs}
\label{motivation}
Consider an arbitrary attention or MLP layer, denoting step indices $1$ and $2$ as previous and current steps, respectively. For attention, define queries ($q_i$), keys ($k_i$), and values ($v_i$); for MLP layers, define input activations ($x_i$), and weight matrices ($W_1$, $W_2$). With a mask $M$ with entries in $\{0, 1\}$ that identifies intermediate activations exhibiting the largest cross-step changes, the outputs at the current step can be approximated by recomputing only the subset of most significantly changing activations. This involves recomputing the columns of $k^T$ or $W_1$ and corresponding rows of $v$ or $W_2$ identified by $M$, while reusing the unchanged activations from the previous step:

{
  \setlength{\abovedisplayskip}{2pt}      
  \setlength{\belowdisplayskip}{2pt}      
  \setlength{\abovedisplayshortskip}{2pt} 
  \setlength{\belowdisplayshortskip}{2pt}

  \begin{align}\label{eqn:delta}
  \text{o}_2^\text{attn} &\approx \text{o}_1^\text{attn} - [\text{softmax}(q_1 \; @ \; k_1^T) * \;M] \; @ \; v_1 \;\,+ [\text{softmax}(q_2 \; @ \; k_2^T) * \;M] \; @ \; v_2\\
 \text{o}_2^\text{mlp} &\approx \text{o}_1^\text{mlp} - [\;\;\;\;\;\;\;\;\;\; \text{gelu}(x_1 \; @ \; W_1) * M] \; @ \; W_2 + [\;\;\;\;\;\;\;\;\; \text{gelu}(x_2 \; @ \; W_1) * M] \; @ \; W_2
  \end{align}
}

\begin{table}[h]
\centering
\caption{$R^2$ scores between true cross-step activation change and approximated change.}
\begin{tabular}{|c|cc|cc|}
\hline
 & \text{Attention Active} & \text{Explained Change} & \text{MLP Active} & \text{Explained Change} \\
\hline
\text{HunyuanVideo} & 5\% & 92.4\% & 25\% & 70.8\% \\
\text{FLUX.1-dev} & 25\% & 90.7\% & 15\% & 69.0\% \\
\hline
\end{tabular}
\label{tab:sparsity_comparison}
\end{table}

The empirical results summarized in Table \ref{tab:sparsity_comparison} show that recomputing just a small fraction of the fastest changing activations captures a substantial portion of the true cross-step activation change. We measure $R^2(\text{o}_2^\text{dense} - \text{o}_1^\text{dense},\; \text{o}_2^\text{approx} - \text{o}_1^\text{dense})$ averaged across steps, layers, tokens, and 25 random prompts.

These findings motivate Chipmunk’s caching approach (Fig. \ref{fig:delta-packing-fusion}, left), which recomputes only the fastest changing vectors in the output linear combinations of attention and MLP layers, while reusing the remaining cached vectors. As a more general motivation beyond this specific context, we analyze the residual structure inherent in multi-step transformer inference more broadly. As shown in Appendix \ref{appx-dit}, the final latent space paths produced by rectified flow DiTs can be decomposed into distinct, individually identifiable scaled vectors output by attention and MLP layers. Thus, the strategy of dynamically allocating more sequential steps to the fastest changing individual vectors can potentially apply to various multi-step transformer settings—such as recurrent transformer architectures \cite{geiping2025scaling} or autoregressive models decoding language \cite{vaswani2023attentionneed} or videos \cite{magi1}.

\subsection{Hardware-Aware Column-Sparse Deltas in Attention and MLPs}\label{hw-efficiency}
Recomputing individual vectors output by attention and MLPs corresponds to [1, 1] unstructured sparsity on the intermediate activations, which is not efficient on modern GPUs. To remedy this, first, we describe how hardware-aware column-sparse recomputation achieves a strong balance of reduced approximation error and hardware efficiency. We then detail optimizations such as voxel-based token reordering and efficient sparse memory operations.

\subsubsection{Tile Packing: Mapping Sparse Computation to a Compacted Dense Computation}
Chipmunk uses column sparse patterns on intermediate activations \cite{chen2021colsparse, Li_2022colsparse,ye2025flashinfer}, where a group of contiguous tokens only activates a certain set of individual keys/values (attention) or neurons (MLPs), to achieve low approximation error while remaining hardware-efficient.

\begin{table}[h]
\centering
\caption{Comparing hardware efficiency and approximation error of block sparsity vs. column sparsity. Approximation error is measured as unexplained variance as described in \ref{motivation}. Runtime is measured as a \% relative to dense computation at HunyuanVideo sequence length (118k).}
\begin{tabular}{|c|ccc|ccc|}
\hline
 & \multicolumn{3}{c|}{Attention} & \multicolumn{3}{c|}{MLP} \\
 \hline
 & Sparsity & Error & Runtime & Sparsity & Error & Runtime \\
\hline
{[192, 128] Block} & 95\% & 17.4\% & 5.1\% & 75\% & 61.3\% & 27.1\% \\
{[192, 1] Column} & 95\% & 8.5\% & 8.5\% & 75\% & 29.1\% & 27.7\%  \\
\hline
\end{tabular}
\label{tab:block-vs-col}
\end{table}

\begin{figure}
    \centering
    \includegraphics[width=0.9\linewidth]{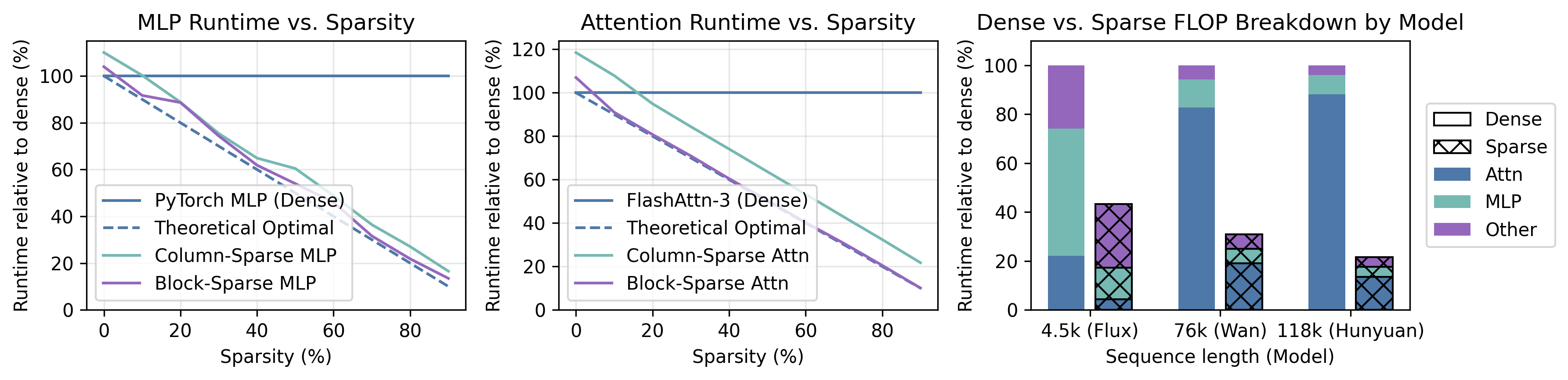}
    \caption{\textit{Left, Middle}: Sparse kernel runtime scales linearly with sparsity. Column sparsity is competitive with block sparsity. \textit{Right}: FLOP breakdown by model at Chipmunk hyperparameters.}
    \label{fig:sparse-kernel-perf}
\end{figure}

\paragraph{Block sparsity is efficient but has a higher approximation error than column sparsity.} As described in Section \ref{background}, GPU kernels achieve peak efficiency by computing large block matrix multiplications. Block sparse kernels maintain hardware-efficiency because the outer loop simply skips certain tiles while leaving inner logic unchanged \cite{tillet2019triton}. Block sparsity achieves near-optimal performance (Fig. \ref{fig:sparse-kernel-perf}, purple lines), but suffers from 2x higher approximation error than finer-grained column sparsity patterns (Table \ref{tab:block-vs-col}, Fig. \ref{fig:attn-mlp-error}).

\paragraph{Column sparsity can be made efficient with sparse gathers from HBM to SRAM.} We implement column-sparse attention and MLP kernels to achieve 2x less approximation error than block sparsity (Fig. \ref{fig:attn-mlp-error}, purple \& cyan lines) while maintaining competitive speedups (Table \ref{tab:block-vs-col}, Fig. \ref{fig:sparse-kernel-perf}). Traditional high-performance dense attention and MLP kernels, such as FlashAttention-3 and cuBLAS GEMMs, use the following structure: (1) load dense 2D tensors from HBM to SRAM, (2) compute the large matrix multiplications with tensor core instructions, and (3) store results back to HBM  \cite{shah2024flashattention3fastaccurateattention,spector2024thunderkittenssimplefastadorable}. To maintain tensor core utilization with column sparsity, we modify step (1) to pack sparse keys/values from non-contiguous rows in global memory into a dense tile in shared memory (Fig. \ref{fig:delta-packing-fusion}) \cite{chen2021colsparse, Li_2022colsparse,ye2025flashinfer}. Tangentially, Chipmunk’s sparse kernels can take on any static sparsity pattern (e.g. Sliding Tile Attention \cite{zhang2025fastvideogenerationsliding}, DiTFastAttn \cite{yuan2024ditfastattnattentioncompressiondiffusion}) by simply passing in a particular set of indices. For low-level kernel optimizations, see Appendix \ref{appx-kernel-opt}.

\paragraph{Column sparsity routes a chunk of tokens to a set of individual key/values (attention) or neurons (MLP).} Given an arbitrary attention or MLP computation, column sparse patterns mask intermediate activations with granularity $[C, 1]$ \cite{chen2021colsparse, Li_2022colsparse,ye2025flashinfer}. This means that, given a chunk of contiguous tokens of size $C$, the chunk only activates a certain set of individual keys/values (attention) or neurons (MLPs). This is in contrast to MoE models, which route individual tokens to chunks of contiguous MLP neurons (experts) \cite{jiang2024mixtralexperts}. Column sparsity is parameterized by the chunk size $C$, and on H100 GPUs, we find $C=192$ is efficient.

\paragraph{Reordering tokens into video voxel/image patch chunks improves attention quality.} By default, contiguous tokens follow raster order (left-to-right, top-down, sequential frames). However, small video voxels (3D cubes of pixels) or image patches have similar color and brightness, and we expect them to exhibit similar interactions with other tokens. Thus, we reorder tokens at the beginning of the diffusion process such that each contiguous token chunk, which is routed to the same set of sparse indices, corresponds to a voxel/patch (Alg. \ref{alg:voxel}). We find this to improve attention approximation quality by 1.2x (Fig. \ref{fig:attn-mlp-error}).

\begin{table}[ht]
\centering
\begin{tabular}{@{}p{0.78\linewidth} c@{}}
\toprule
\textbf{Method} & \textbf{Speedup} \\
\midrule
Attention \& unfused column-sum
& 1.0x \\
$\rightarrow$ Fused Kernel                           & 3.26x \\
\midrule
MLP \& unfused cache operations                                                        & 1.0x \\
$\rightarrow$ Fused Kernel
& 1.41x \\
\midrule
Attention mask to indices conversion (unfused) & 1.0x \\
$\rightarrow$ Fused Kernel
& 1.39x \\
\midrule
PyTorch top-k & 1.0x \\
$\rightarrow$ Approximate top-k & 3.42x \\
\bottomrule
\end{tabular}
\caption{Kernel optimizations applied to Chipmunk's end-to-end algorithm measured on FLUX.1-dev shapes on H100-SXM5 GPUs with CUDA 12.8 and PyTorch 2.5.0.}
\label{tab:pattern-identification}
\end{table}
\subsection{Chipmunk}

Across a full DiT generation, Chipmunk accelerates Diffusion Transformer (DiT) inference by interleaving dense and sparse-delta steps to exploit the slow changing activations across steps, and natively stacks with coarser forms of caching (e.g., per-step \cite{liu2024timestep}, per-layer \cite{wimbauer2024cachecanacceleratingdiffusion}, per-token \cite{zou2025acceleratingdiffusiontransformerstokenwise}). Dense steps refresh cached activations and identify dynamic sparsity patterns, while sparse steps efficiently compute sparse activation deltas against cached activations. Structured column-chunk sparsity patterns are applied to the intermediate activations of both attention and MLP to enable hardware-efficient sparse algorithms. To improve this column-sparse approximation quality, Chipmunk applies a reordering to tokens at the beginning of the diffusion process such that each contiguous chunk of $c$ tokens corresponds to a 3D video voxel (Alg. \ref{alg:voxel}). All tokens in one voxel will therefore share the same sparsity pattern in later steps. The inverse reordering is then applied before the final decoding to pixel space. Optimized fused kernels minimize the overhead of computing dynamic sparse activation deltas, and their performance impacts are summarized in Table \ref{tab:pattern-identification}. In the discussion of this section, we describe a reference-correct implementation of Chipmunk, with kernel optimizations further detailed in Appendix \ref{appx-kernel-opt}.

\subsubsection{Chipmunk Attention}

\paragraph{Dense Steps.}
We run full scaled-dot-product attention on the reordered sequence

$$
\mathbf Q,\mathbf K,\mathbf V\in\mathbb R^{B\times H\times N\times E},
\qquad
\mathbf P=\operatorname{softmax}\!\bigl(\tfrac{\mathbf Q\mathbf K^{\!\top}}{\sqrt E}\bigr),\;
\mathbf O_{\text{dense}}=\mathbf P\mathbf V .
$$

We then partition the query axis of attention matrix $\mathbf P$ into chunks of $c$ contiguous tokens (the queries of one voxel),

$$
\mathbf P\rightarrow
\text{reshape}\bigl(B,H,\tfrac{N}{c},c,N\bigr)
$$

and sum over the $c$ dimension. The result is a “column-sum” tensor

$$
\mathbf D\in\mathbb R^{B\times H\times \frac{N}{c}\times N},\qquad 
\mathbf D_{b,h,i,j}=\!\!\sum_{q=ic}^{(i+1)c-1}\!P_{b,h,q,j}
$$

which tells us, for each voxel $i$, how much total attention probability it assigns to key $j$ (Alg. \ref{alg:colsumattention}). Computing full attention and the column-sum are fused into a single kernel (Table \ref{tab:pattern-identification} Row Group 1, Appendix \ref{appx-colsum}).

A top-k over the last dimension then selects the most-attended keys/values for each query voxel:
\[
\;
\text{idx}_{b,h,i,:k}
  =\operatorname{TopK}_k\!\bigl(\mathbf D_{b,h,i,:}\bigr)
\;
\]
These indices are cached for the upcoming sparse steps. Finally, Chipmunk defines the attention activation cache as

$$
\;
\mathbf O_{\text{cache}}
   = \mathbf O_{\text{dense}}
   - \text{softmax} \bigl(\mathbf Q\,\mathbf K^\top_{\!\text{idx}}\bigr)\,\mathbf V_{\text{idx}}
\;
$$

such that subsequent steps can directly add to this cache to compute a sparse replacement of the top attention interactions identified by the cached indices (Alg. \ref{alg:chipmunkattention}).

\paragraph{Sparse Delta Steps.}
In subsequent sparse steps Chipmunk only recomputes the top attention interactions defined by the indices cache (Alg. \ref{alg:colsparsedeltaattn}):

$$
\Delta\mathbf O
   = \text{softmax} \bigl(\mathbf Q\,\mathbf K^\top_{\!\text{idx}}\bigr)\,\mathbf V_{\text{idx}},
\qquad
\mathbf O = \mathbf O_{\text{cache}} + \Delta\mathbf O .
$$

The computational complexity of attention thus drops from $\mathcal O(N^2)$ to $\mathcal O(Nk)$, reproducing the dense result when $k=N$. Conceptually, this computation corresponds exactly to Fig. \ref{fig:delta-packing-fusion} (Left), where we visualize the sparse recomputation of the fastest changing vectors in attention's output linear combination (scaled rows of $\mathbf V$), while reusing the remaining vectors from the cache.

\subsubsection{Chipmunk MLP}
For MLP layers, Chipmunk similarly alternates dense and sparse-delta computations. Dense steps fully compute activations and initialize caches, while sparse steps efficiently identify and recompute only activations that significantly change across steps (Alg. \ref{alg:chipmunkmlp}).

\paragraph{Dense steps.}
A standard two-layer MLP layer produces

$$
\mathbf A = \sigma\!\bigl(\mathbf X\mathbf W_1^{\!\top}+\mathbf b_1\bigr),\;
\mathbf O_{\text{dense}} = \mathbf A\mathbf W_2 + \mathbf b_2 ,
$$

with shapes
$\mathbf X\in\mathbb R^{B\times N\times D},\;
 \mathbf W_{1,2}\in\mathbb R^{F\times D}$.
Chipmunk caches three tensors on dense steps:
\begin{enumerate}
\item{Tile-mean pre-activation
   $\mathbf T_{\!m} = \text{mean}_c(\text{reshape}_c(\mathbf X)\mathbf W_1^{\!\top}) \in\mathbb R^{B\times N/c\times F}$.}
\item{Full activation tensor $\mathbf A_{\text{cache}}$.}
\item{MLP output $\mathbf M_{\text{cache}} = \mathbf O_{\text{dense}}$.}
\end{enumerate}

\paragraph{Sparse Delta Steps.}
Chipmunk uses a token-merged approximation to identify the $k$ neurons that have changed the most since the last step (for each size-$c$ chunk of contiguous tokens) (Alg. \ref{alg:computemlpindices}):

$$
\Delta_{\text{tm}}=\lvert\mathbf T - \mathbf T_{\!m}\rvert,\quad
\text{idx} =\operatorname{TopK}(\Delta_{\text{tm}}) \in\mathbb R^{B\times N/c\times k},
$$

We then recompute only those token-neuron interactions and reuse the rest from the cache (Alg. \ref{alg:colsparsedeltamlp}). As an optimization, we compute MLP in a single step—without the separate subtraction and addition of Eq. \ref{eqn:delta}—because of the static nature of the second operand ($W_2$) (Appendix \ref{appx-mlp-opt}). 

$$
\Delta\mathbf M
   = \sigma\bigl(\mathbf X\,\mathbf W_{1,\text{idx}}^\top - \mathbf A_{\text{cache, idx}}\bigr)\mathbf W_{2,\text{idx}},
\qquad
\mathbf O = \mathbf M_{\text{cache}} + \Delta\mathbf M .
$$

Analogous to attention, this computation corresponds exactly to Fig. \ref{fig:delta-packing-fusion}, Left. We compute a sparse replacement of the fastest changing vectors in MLP's output linear combination (scaled rows of $\mathbf W_2$), while reusing the remaining vectors from the cache.

\subsubsection{Natively stacking with coarser DiT caching techniques.} Chipmunk's caching and sparsity operate at a fine granularity within attention and MLP layers, naturally complementing other acceleration methods that cache at coarser granularities such as steps, layers, or tokens. To demonstrate this, in the following section, we stack Chipmunk with step caching and sliding tile attention to achieve a strong speed-quality tradeoff.
\section{Experiments}\label{experiments}
\begin{figure}[htbp]
    \centering
    \includegraphics[width=0.45875\linewidth]{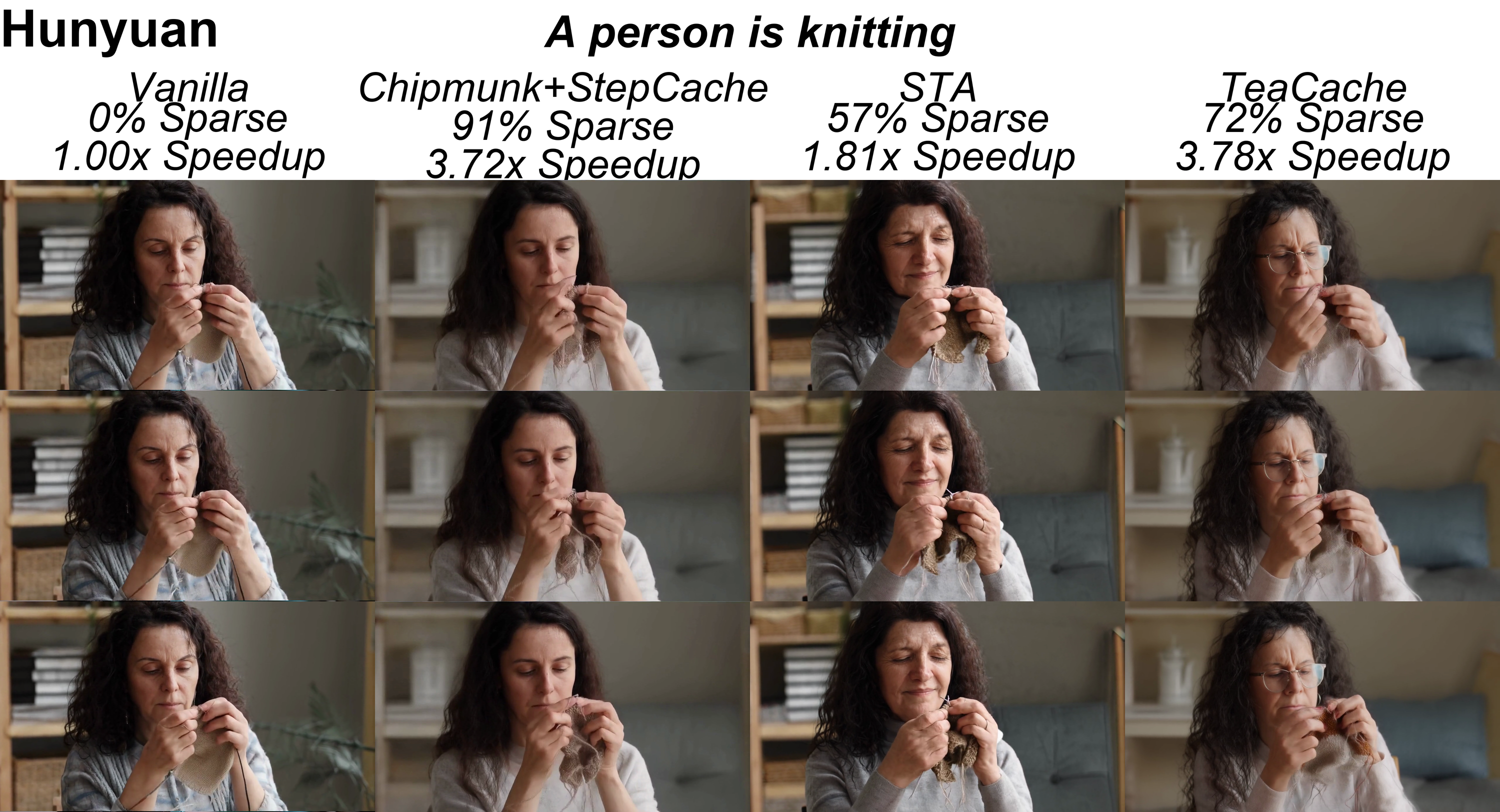}
    \hfill
    \includegraphics[width=0.52375\linewidth]{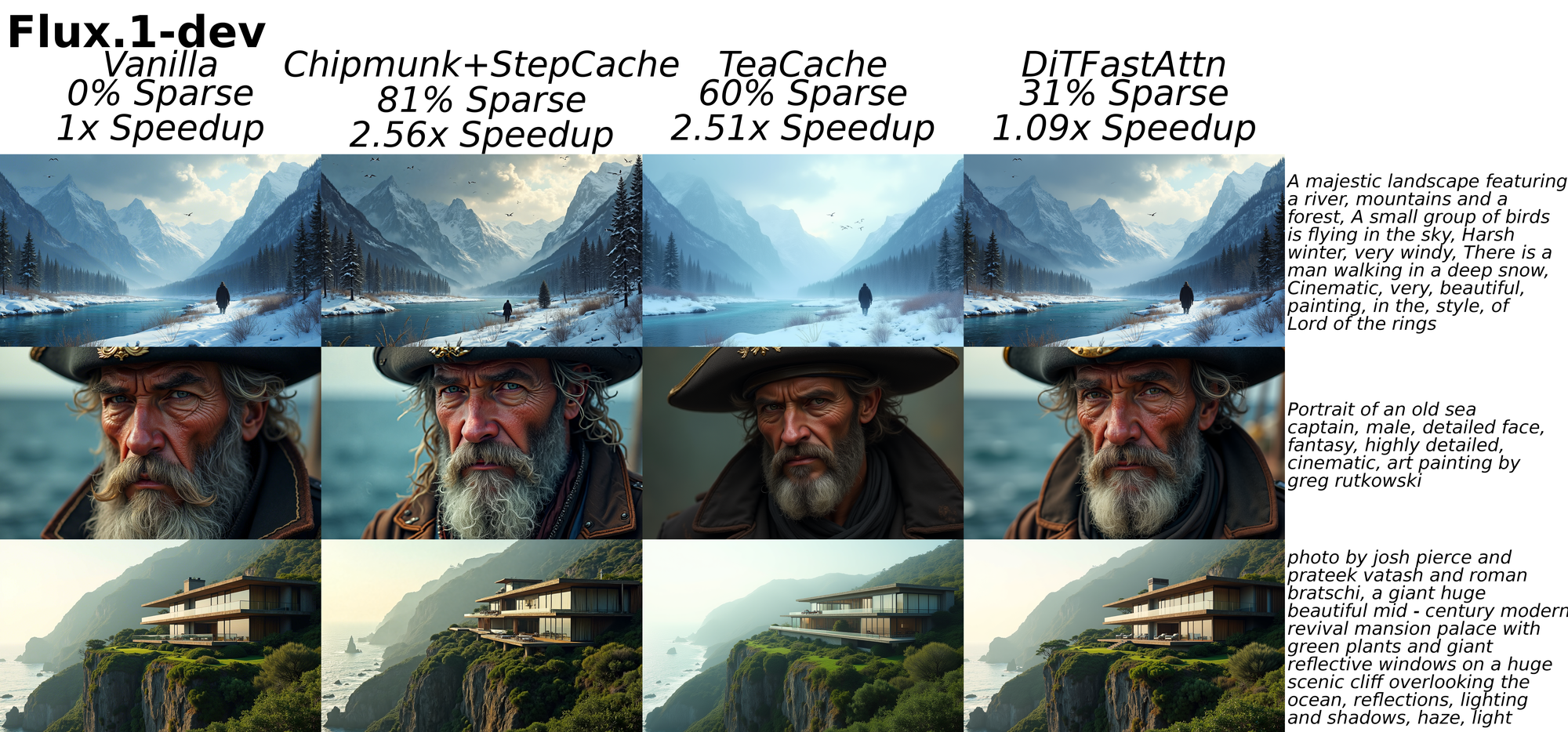}
    
    \caption{Qualitative comparisons across videos (left) and images (right). For videos, frames are stacked vertically (down is later). See Appendix \ref{appx-evaluations} for more.}
    \label{fig:mini-qualitative-grid}
\end{figure}

We evaluate Chipmunk against state-of-the-art DiT acceleration methods on text-to-video and text-to-image tasks. First, we outline our experimental setup (\ref{exp-setup}). Quantitative comparisons in Tables \ref{tab:performance_comparison} and \ref{tab:performance_comparison_img} demonstrate Chipmunk’s efficiency and quality improvements (\ref{exp-quantitative}). Qualitative examples (Fig. \ref{fig:mini-qualitative-grid}) illustrate Chipmunk’s preservation of visual quality under significant acceleration (\ref{exp-qualitative}).

\begin{table}[!ht]
\renewcommand{\arraystretch}{1} 
\centering
\begin{tabular}{>
{\raggedright\arraybackslash}m{3.8cm}|cc|>{\raggedright\arraybackslash}m{1.2cm}>{\raggedright\arraybackslash}m{1.3cm}>{\raggedright\arraybackslash}m{1.65cm}}
\hline
\multirow{2}{*}{\textbf{Method}} & \multicolumn{2}{c|}{\textbf{Efficiency}} & \multicolumn{3}{c}{\textbf{VBench Dimension}} \\
\cline{2-6}
 & Speedup $\uparrow$ & Latency (s) $\downarrow$ & Total $\uparrow$ & Quality $\uparrow$ & Semantic $\uparrow$ \\
\hline\hline
\multicolumn{6}{c}{HunyuanVideo, $T$=50 (720 x 1280 x 129)} \\
\hline
Hunyuan  & 1x & 1030s & 83.24 & 85.09 & 75.82 \\
\hline
STA & 1.79x & 575s & 82.46 & \textbf{84.63} & 73.83 \\
\rowcolor{gray!15}
Chipmunk & \textbf{2.16x} & \textbf{477s} & \textbf{82.94} & 84.60 & \textbf{76.3} \\
\hline
Step Caching (TeaCache) & 3.69x & 279s & 80.79 & 82.87 & 72.5 \\
\rowcolor{gray!15}
Chipmunk+Step Cache & \textbf{3.72x} & \textbf{277s} & \textbf{82.5} & \textbf{84.23} & \textbf{75.6} \\
\hline\hline
\multicolumn{6}{c}{WAN2.1, $T$=50 (720 x 1280 x 121)} \\
\hline
WAN2.1  & 1x & 1357s & 81.47 & 83.57 & 73.08 \\
\hline
STA & 1.36x & 998s & \textbf{81.84} & \textbf{83.65} & \textbf{74.60} \\
\rowcolor{gray!15}
Chipmunk+STA & \textbf{1.56x} & \textbf{870s} & 81.71 & 83.61 & 74.12 \\
\hline
Step Caching (TeaCache) & 2.0x & 678s & 81.17 & 83.24 & 72.87 \\
\rowcolor{gray!15}
Chipmunk-56\% +STA+Step Cache & 2.20x & 616s & \textbf{81.73} & \textbf{83.74} & 73.69 \\
\rowcolor{gray!15}
Chipmunk-73\% +STA+Step Cache & \textbf{2.67x} & \textbf{508s} & 81.11 & 82.88 & \textbf{74.05} \\
\hline\hline
\end{tabular}
\caption{Performance comparison of various methods across different datasets for video generation. \textit{Note}: Chipmunk-$X$\% denotes a sparsity level of $X$\% to assess the speed-quality tradeoff.}
\label{tab:performance_comparison}
\end{table}

\begin{table}[!ht]
\renewcommand{\arraystretch}{1} 
\centering
\begin{tabular}{>{\raggedright\arraybackslash}m{4cm}|ccc|c}
\hline
\multirow{2}{*}{\textbf{Method}} & \multicolumn{3}{c|}{\textbf{Efficiency}} & \multicolumn{1}{c}{\textbf{Visual Quality}} \\
\cline{2-5}
 & FLOPs $\downarrow$ & Speedup $\uparrow$ & Latency (s) $\downarrow$ & ImRe $\uparrow$ \\
\hline\hline
\multicolumn{5}{c}{FLUX.1-dev, $T$=50 (768 × 1280)} \\
\hline
Flux & 100\% & 1x & 6.60s & 0.76\\
\hline
STA         & 84\% & 1.15x & 5.73s  & 0.75 \\
DiTFastAttn  & 83\% & 1.09x & 6.05s & \textbf{0.80} \\
\rowcolor{gray!15}
Chipmunk & \textbf{58\%} & \textbf{1.41x} & \textbf{4.90s} & \textbf{0.80} \\
\hline
Step+Token Caching (ToCa)     & 66\% & 1.51x & 4.37s & 0.76 \\
Step Caching (TeaCache) & 39\% & 2.51x & 2.64s & 0.68 \\
\rowcolor{gray!15}
Chipmunk+Step Cache & \textbf{31\%} & \textbf{2.56x} & \textbf{2.57s} & \textbf{0.77} \\
\hline\hline
\end{tabular}
\caption{Performance comparison of various methods for ImageReward for image generation.}
\label{tab:performance_comparison_img}
\end{table}

\begin{table}[!ht]
\renewcommand{\arraystretch}{1} 
\centering
\begin{tabular}{>{\raggedright\arraybackslash}m{4cm}|ccc|cc}
\hline
\multirow{2}{*}{\textbf{Method}} & \multicolumn{3}{c|}{\textbf{Efficiency}} & \multicolumn{2}{c}{\textbf{Visual Quality}} \\
\cline{2-6}
 & FLOPs $\downarrow$ & Speedup $\uparrow$ & Latency (s) $\downarrow$ & GenEval $\uparrow$ & CLIP $\uparrow$ \\
\hline\hline
\multicolumn{6}{c}{FLUX.1-dev, $T$=50 (768 × 1280)} \\
\hline
Flux & 100\% & 1x & 6.60s & 0.66 & 31.07 \\
\hline
Step+Token Caching (ToCa)     & 66\% & 1.51x & 4.37s & 0.65 & 31.21 \\
Step Caching (TeaCache) & 45\% & 2.23x & 2.95s & 0.61 & 31.37 \\
\rowcolor{gray!15}
Chipmunk-77\%+Step Cache & \textbf{31}\% & \textbf{2.56x} & \textbf{2.57s} & 0.62 & 31.18 \\
\rowcolor{gray!15}
Chipmunk-65\%+Step Cache & {38\%} & {2.25x} & {2.93s} & {\textbf{0.66}} & {\textbf{31.43}} \\
\hline\hline
\end{tabular}
\caption{Performance comparison of various methods across various GenEval and CLIP. \textit{Note}: Chipmunk-$X$ denotes a sparsity level of $X$\% to assess the speed-quality tradeoff.}
\label{tab:performance_comparison_img_genevalclip}
\end{table}

\subsection{Setup}\label{exp-setup}
\paragraph{Models.} We evaluate Chipmunk across three state-of-art DiT models: HunyuanVideo and WAN2.1 for text-to-video generation and FLUX.1-dev for text-to-image generation, all using their default number of generation steps (50). As shown in Fig. \ref{fig:sparse-kernel-perf} (right), these models vary in sequence and FLOP breakdown, creating a strong evaluation of acceleration ability in different inference regimes. Of the three models, FLUX.1-dev has the smallest sequence length at 4.5k and allocates a majority of FLOPs to MLP layers. WAN2.1 and HunyuanVideo are extremely bound by attention at sequence lengths of 76k and 118k, respectively. All benchmarks are performed on H100-SXM5 GPUs with CUDA 12.8 and PyTorch 2.5.0, comparing against a 650 TFLOP FlashAttention-3 baseline \cite{shah2024flashattention3fastaccurateattention}.

\paragraph{Baselines.} We compare Chipmunk against a number of recent DiT acceleration techniques including TeaCache \cite{liu2024timestep}, ToCa \cite{zou2025acceleratingdiffusiontransformerstokenwise}, Sliding Tile Attention (STA) \cite{zhang2025fastvideogenerationsliding}, and DiTFastAttn \cite{yuan2024ditfastattnattentioncompressiondiffusion}. TeaCache is a method for dynamically reusing DiT step outputs during slow-changing portions of the generation process, and ToCa dynamically computes token importance scores to recompute activations for only the most important tokens while reusing the rest. STA extends sliding window attention to 3D video to exploit the natural locality in attention computations, and DiTFastAttn reuses non-local attention interactions in a subset of steps to avoid a full recomputation. Here, we also stack Chipmunk with a simple static step cache schedule, untuned for any particular model/prompt. Full baseline descriptions are available in Appendix \ref{appx-hyperparameters}.

\paragraph{Evaluations and Metrics.} We evaluate text-to-video generation with the standard VBench evaluation, comprised of 16 dimensions that are weighted to produce a final score shown to align with human judgement \cite{huang2023vbenchcomprehensivebenchmarksuite}. For text-to-image generation, we compute three standard metrics: (i) ImageReward (ImRe) \cite{xu2023imagereward}, which is a common human-preference trained reward model; (ii) CLIP, which is a widely used metric assessing semantic alignment between text prompts and images (evaluated on a random subset of 10,000 images from MSCOCO2017) \cite{radford2021learningtransferablevisualmodels,lin2015microsoftcococommonobjects}; and (iii) GenEval, which is another automatic evaluation metric designed specifically to gauge the quality and accuracy of image generations by comparing them to reference textual prompts, capturing aspects such as relevance, coherence, and descriptive fidelity \cite{ghosh2023genevalobjectfocusedframeworkevaluating}.

\paragraph{Hyperparameters.} In a warm-up phase for each model, we test 100 E2E generations in order to select values of \textit{MLP sparsity level} and \textit{attention sparsity level} that will achieve 95\% of the change in activations across steps. For the \textit{step schedule} hyperparameter, we choose a simple schedule of interleaving 1 dense step every 10 sparse steps; further optimization could yield additional efficiency improvements.

\subsection{Quantitative Results}\label{exp-quantitative}
Across image and video models, Chipmunk achieves the largest acceleration levels while maintaining near-lossless quality across ImageReward, CLIP, and VBench (Tables \ref{tab:performance_comparison}, \ref{tab:performance_comparison_img}, and \ref{tab:performance_comparison_img_genevalclip}). We evaluate results in the context of three regimes: \textbf{(1) Acceleration Held Constant. }Holding acceleration level constant, Chipmunk's fine-grained caching of individual attention and MLP vectors enable it to maintain higher quality than TeaCache, which caches full step outputs (HunyuanVideo, rows 4-5; Flux Table \ref{tab:performance_comparison_img}, rows 6-7; Flux Table \ref{tab:performance_comparison_img_genevalclip}, rows 3-4). \textbf{(2) Quality Held Constant. }When quality is held nearly constant, Chipmunk achieves higher acceleration than STA, ToCa, and DiTFastAttn, due to its dynamic, granular identification of sparsity patterns (HunyuanVideo, rows 2-3; WAN2.1, rows 2-3; Flux Table \ref{tab:performance_comparison_img}, rows 2-5; Flux Table \ref{tab:performance_comparison_img_genevalclip}, rows 2 and 4). \textbf{(3) Chipmunk Speed-Quality Tradeoff. }Chipmunk's sparsity hyperparameter can modulate the speed-quality tradeoff (WAN2.1, rows 5-6, Flux Table \ref{tab:performance_comparison_img_genevalclip}, rows 4-5).

\subsection{Qualitative Results}\label{exp-qualitative}

In videos, we observe that Chipmunk preserves small, high-moving parts such as hands knitting (Fig. \ref{fig:mini-qualitative-grid}). In images, Chipmunk maintains strong visual quality, including details from long prompts. In both text-to-image and text-to-video tasks, Chipmunk enhances details on the subject of the generation, such as the buttons on the captain's jacket. We hypothesize this is due to the $\sim$90\% sparse attention matrix focusing on the most relevant parts of the prompt (e.g., the subject). We speculate Chipmunk’s improved ImageReward, VBench, and CLIP scores may stem from this focus on subject prominence, as these metrics evaluate how closely outputs align with prompts that largely describe a subject \cite{radford2021learningtransferablevisualmodels,xu2023imagereward}. For a sample of playable VBench video generations, please see these anonymized YouTube links: \href{https://www.youtube.com/watch?v=etquKck_wtc}{WAN2.1} and \href{https://www.youtube.com/watch?v=rr0Pg4LHqVI}{HunyuanVideo}.

\textbf{Failure Modes.} At times, the background of videos appears slightly out of focus (bookshelf behind the woman knitting), which we similarly speculate can be attributed to the sparsity in the attention matrix concentrating on subject-based parts of the prompt. In text-to-image tasks, even though Chipmunk maintains prompt adherence and high visual quality, we find minor differences in details of the generations when compared to the reference images (e.g., the number of background birds in row 1), likely due to the number of FLOPs removed.

\section{Related Work}

\textbf{Step Distillation and Consistency Models.} Step distillation and related methods such as Consistency Models reduce inference time by training few-step models to approximate the outputs of a many-step teacher models\cite{song2023consistencymodels, esser2024scalingrectifiedflowtransformers,zhang2024acceleratingdiffusionmodelsonetomany}. These approaches uniformly allocate fewer inference steps across all model computations, effectively learning to compensate by proportionally increasing the magnitude of latent-space updates per step. In contrast, Chipmunk dynamically allocates computational resources within a step, \textit{selectively recomputing} the fastest-changing activation vectors without training. 

\textbf{Training-Free Activation Caching. }Training-free activation caching methods can be categorized into three buckets depending on their granularity: (i) at the step level, (ii) at the layer level, and (iii) at the token level \cite{zou2024acceleratingdiffusiontransformersdual}. Approaches such as TeaCache \cite{liu2024timestep} and AdaCache \cite{kahatapitiya2024adaptivecachingfastervideo} operate at the per-step level, using adaptive mechanisms to determine optimal steps to skip. Approaches such as BlockCache \cite{wimbauer2024cachecanacceleratingdiffusion} and $\Delta$-DiT \cite{chen2024deltadittrainingfreeaccelerationmethod} operate at the per-layer attention/MLP layer level. Finally, approaches like ToCa and DaTo \cite{zou2025acceleratingdiffusiontransformerstokenwise,zhang2024tokenpruningcachingbetter} operate at the per-token level within attention and MLP layers. Chipmunk has finer granularity than the aforementioned methods, operating at the per-vector level within attention and MLP layers. Therefore, it is possible to complement Chipmunk with any of these approaches.

\textbf{Sparse Attention Computation. }The natural sparsity and locality in large attention matrices is a common observation , and many methods exploit it \cite{zhang2025sageattentionaccurate8bitattention,tan2025dsvexploitingdynamicsparsity,xu2025xattentionblocksparseattention,tang2024questqueryawaresparsityefficient,chen2024longloraefficientfinetuninglongcontext}. Sliding Tile Attention and Sparse VideoGen implement static attention masks to only attend to spatiotemporally local tokens \cite{zhang2025fastvideogenerationsliding,xi2025sparsevideogenacceleratingvideo}. Similarly, DiTFastAttention implements a sliding window pattern to recompute static local attention \cite{yuan2024ditfastattnattentioncompressiondiffusion}. Chipmunk can be complemented with any static sparsity pattern by passing the relevant indices as kernel inputs.
\section{Conclusion}

We introduce Chipmunk, a novel, training-free approach designed to accelerate DiT inference by dynamically exploiting the inherent sparsity and slow-changing patterns of intermediate activations. Chipmunk leverages the shared computational structure of attention and MLPs to identify and recompute only the activations that change most rapidly, caching the rest for reuse across inference steps. Chipmunk achieves substantial reductions in computational load, resulting in faster inference times with minimal compromises to visual quality.

\textbf{Limitations.} Chipmunk provides optimal speedups primarily for large, compute-bound DiTs, and is less effective for smaller models or those with compact matrix shapes (e.g., Facebook DiT, PixArt-$\alpha$), where sparsity overheads may dominate computational gains. Additionally, while Chipmunk can be applied to models with any number of steps, few-step models with larger activation changes across steps may present less opportunities for caching, resulting in more modest acceleration benefits.

\textbf{Future Work.} Chipmunk is currently training-free, but future work should explore the potential of integrating sparse recomputation into the training process to allow models to learn to dynamically apply sequential steps to the fastest-changing vectors. Separately, in this work, we only stack Chipmunk with per-step caching, but future work may stack it with per-layer and per-token caching as well.

\section{Acknowledgements}
We thank Aaryan Singhal, Rahul Chalamala, Neil Movva, and Hayden Prairie for helpful feedback and discussions during this work.

\bibliographystyle{abbrvnat}   
\bibliography{references}      

\newpage
\appendix

\section{Extended Discussion of Latent Space Path Decompositions}\label{appx-dit}
\begin{figure}
    \centering
    \includegraphics[width=0.8\linewidth]{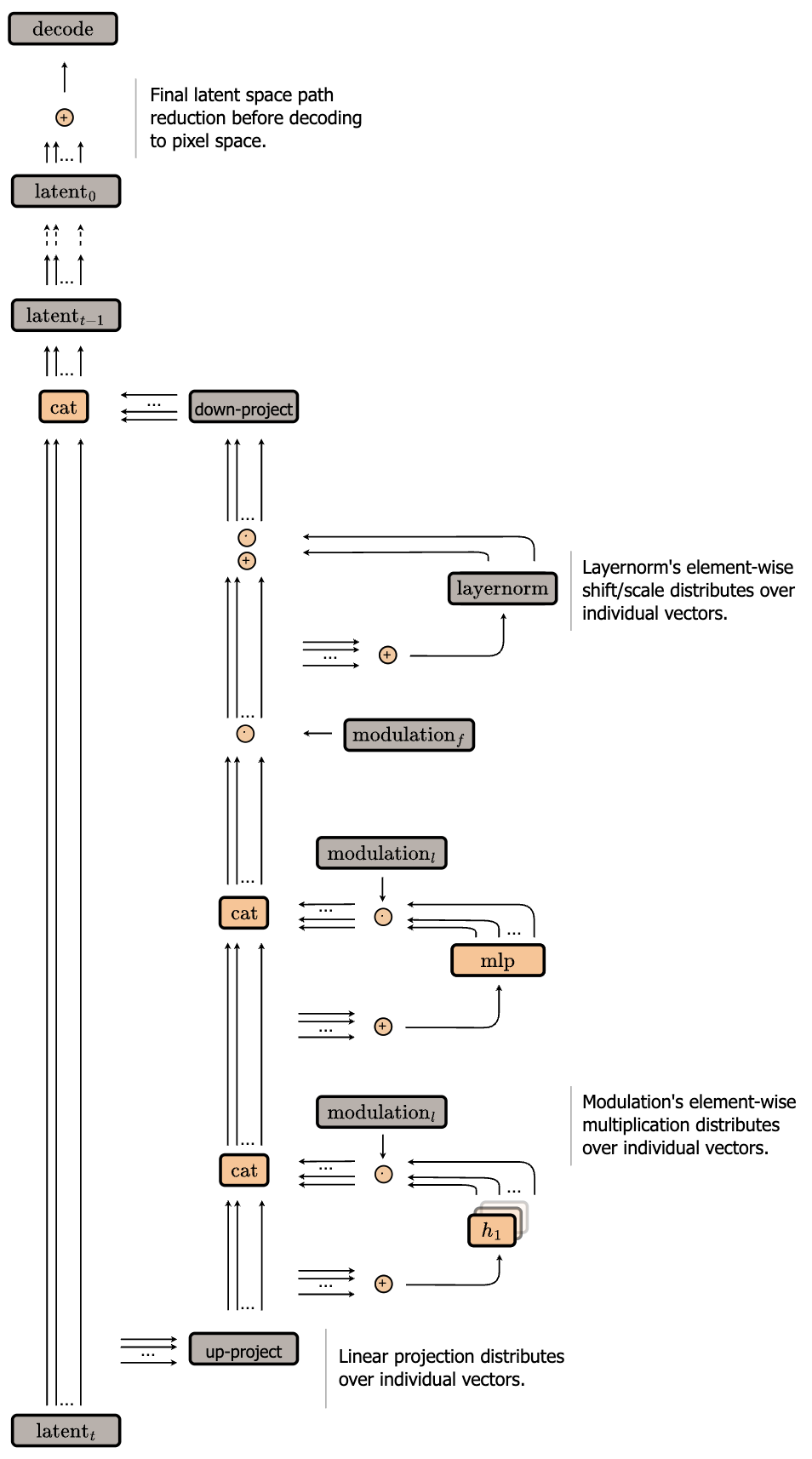}
    \caption{An individual-vector interpetation of the full HunyuanVideo DiT architecture. Attention and MLPs add new vectors to the residual stream, with downstream modulation, layernorm, and linear projection operations distributing over the vector sums.}
    \label{fig:full-dit}
\end{figure}

In this appendix, we formally demonstrate that the latent space paths of rectified flow Diffusion Transformers (DiTs) can be decomposed into a sum of individually scaled vectors output by the attention and MLP layers (Fig. \ref{fig:full-dit}). This decomposition does not claim that each individual vector evolves along an independent path (since the vectors interact via non-linearities) -- rather, we claim that the final latent space path comprises only scaled and shifted vectors output by attention and MLP layers. We assume that in each step of the diffusion process, the DiT output is scaled according to the noise schedule and added to the latent vector representation, as in the architectures of HunyuanVideo and FLUX.1-dev \cite{esser2024scalingrectifiedflowtransformers}.

\paragraph{Attention and MLP Outputs as Scaled Vector Additions.}

\begin{figure}
    \centering
    \includegraphics[width=1\linewidth]{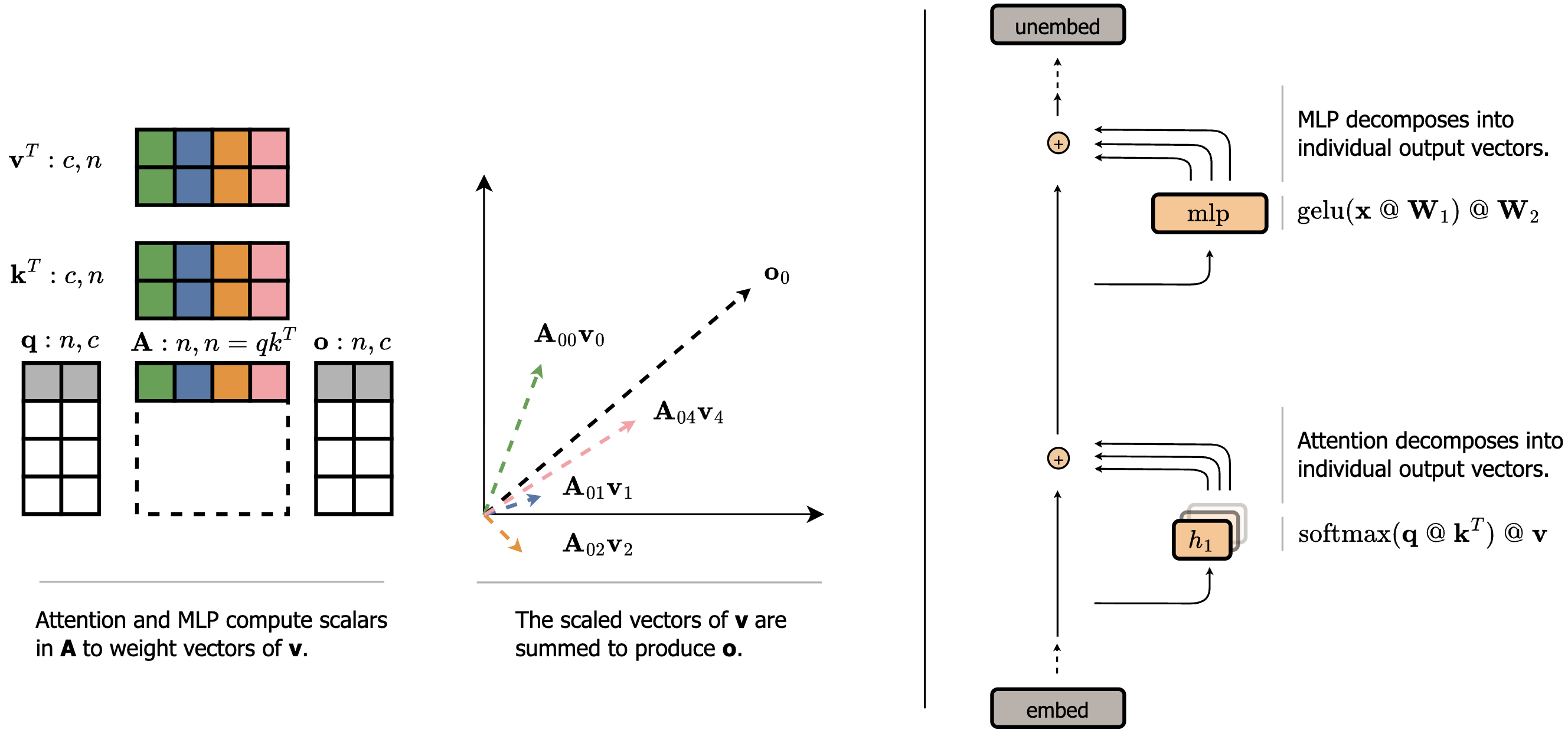}
    \caption{Left: Both MLP and attention operations use a non-linearity to compute the scalar coefficients for a linear combination of value vectors. In attention, the value vectors are dynamic (V is projected from the current token representation). In MLP, the value vectors are static (rows of the weights W2). Right: DiTs accumulate sums of these scaled and shifted value vectors (MLP and attention outputs).}
    \label{fig:attn-mlp-resid}
\end{figure}

We begin by explicitly defining attention and MLP layers within a DiT residual block.

\textbf{Attention}: Given query $q \in \mathbb{R}^{n \times d_c}$, key $k \in \mathbb{R}^{n \times d_c}$, and value $v \in \mathbb{R}^{n \times d_c}$ matrices projected from the residual stream $x \in \mathbb{R}^{n \times d}$, attention computes:
\begin{align}
\text{Attention}(q, k, v) = \text{softmax}\left(\frac{q k^\top}{\sqrt{d_c}}\right) v
\end{align}

To explicitly show the decomposition into individual vectors, we rewrite attention per-head and per-token. Consider a single attention head $h$ with projection matrices $W_q^h, W_k^h, W_v^h \in \mathbb{R}^{d \times d_c}$, and output projection $W_o^h \in \mathbb{R}^{d_c \times d}$. For each token $i$, attention can be expressed as a linear combination of dynamically computed vectors from the combined value-output projection:
\begin{align}
\text{Attn}_h(x_i) &= \sum_j \alpha_{ij}^h \cdot (x_j W_v^h W_o^h),\\
\text{where}\quad \alpha_{ij}^h &= \text{softmax}_j\left(\frac{(x_i W_q^h)(x W_k^h)^\top}{\sqrt{d_c}}\right)
\end{align}

This shows that attention produces a weighted sum of vectors $(x_j W_v^h W_o^h)$, with scalars $\alpha_{ij}^h$ computed via softmax (Fig. \ref{fig:attn-mlp-resid}, left). Each individual vector is thus scaled dynamically based on token interactions.

\textbf{MLP}: Similarly, given weight matrices $W_1 \in \mathbb{R}^{d \times d_{f}}$, $W_2 \in \mathbb{R}^{d_{f} \times d}$ and biases $b_1 \in \mathbb{R}^{d_{f}}$, $b_2 \in \mathbb{R}^{d}$, the MLP operation computes:
\begin{align}
\text{MLP}(x) = \text{GELU}(x W_1 + b_1) W_2 + b_2
\end{align}

The MLP can be viewed as outputting scaled rows of $W_2$ (the bias is a single extra static vector). Specifically, for token $x_i$:
\begin{align}
\text{MLP}(x_i) = \sum_{m=1}^{d_{f}} \text{GELU}(x_i W_1^{:,m} + b_1) \cdot W_2^{m,:} + b_2
\end{align}

Thus, MLP outputs scaled rows from $W_2$, with scalar coefficients computed as GELU applied to the dot product of $x_i$ with each column of $W_1$ (plus a static shift from the bias).

\paragraph{Distributive Operations.}
Having established attention and MLP as explicitly adding scaled vectors, we now formally discuss operations distributing over vector addition within DiTs.

\textbf{Modulations:} Element-wise multiplication by modulation factors $m \in \mathbb{R}^{d}$ distributes trivially:
  \begin{align}
  m \odot \sum_i v_i = \sum_i (m \odot v_i)
  \end{align}

\textbf{Layernorm:} For vectors $v_i$, layer normalization is defined element-wise:
  \begin{align}
  \text{LN}\left(\sum_i v_i\right) = \sum_i\frac{ v_i - \mathbb{E}[\sum_i v_i]}{\sqrt{\text{Var}[\sum_i v_i]}} \gamma + \beta
  \end{align}
  The mean and variance computations do not distribute over vector addition -- however, the resulting modification to the residual stream, as visualized in Fig. \ref{fig:full-dit}, only uses the mean and variance scalars as scale and shift factors, which do distribute over the sum of individual vectors. Thus, in the residual stream framework \cite{elhage2021mathematical}, layernorm can be interpreted as performing a two phase computation: (1) a non-distributive "read" of the residual stream to compute scale and shift factors, (2) a distributive "write" over all vectors in the residual stream using the now fixed scale and shift factors computed in phase (1). This means that after layernorm, the residual stream is still a sum of individually identifiable vectors.
  
\textbf{Linear Projections:} Linear projections defined by matrices $W$ distribute linearly:
  \begin{align}
  W \left(\sum_i v_i\right) = \sum_i W v_i
  \end{align}

\paragraph{Non-Distributive Operations.}
Operations such as softmax and GELU do not distribute over vector addition, but similar to layernorm, they do not directly modify the vectors in the residual stream in a non-distributive manner. Instead, they compute scalar coefficients that scale the vectors newly added to the residual stream (Fig. \ref{fig:full-dit}).

\paragraph{Final Latent Space Path Decomposition.}
Considering these properties collectively, we assert the following:

\begin{enumerate}
\item Attention and MLP layers output sums of scaled vectors, incrementally updating the residual stream.
\item Modulations, scale and shift factors computed by layer normalization, and linear projections distribute over these vector additions.
\item Non-distributive operations (softmax, GELU) are solely used to compute scalar coefficients of newly added vectors without directly modifying the residual stream.
\end{enumerate}

Thus, the latent space path of a DiT output at any inference step $t$, denoted $z_t$, can be formally expressed as a sum of individually scaled vectors from attention and MLP layers across all previous steps $t' > t$ (where $z_0$ is the fully denoised output):
\begin{align}
z_{t} = \text{LN}^t&(\text{Mod}^{t, \text{final}} \odot z_{t + 1}W_e)W_u * s_t \\&+ \sum_{l \in \text{layers}}  \sum_{h}\sum_j \text{LN}^t( \text{Mod}^{l, t,\text{attn}} \odot \alpha_{j}^{h,l,t} (x_j^{l, t}W_v^{h,l}W_o^{h, l}))W_u * s_t \\&+ \sum_{l \in \text{layers}} \sum_m \text{LN}^t(\text{Mod}^{l, t,\text{mlp}} \odot \beta_{m}^{l,t} W_2^{m, :, l})W_u * s_t
\end{align}

Here, vectors $x_j^{l, t}W_v^{h,l}W_o^{h, l}$ and $W_2^{m, :, l}$ represent individual vectors produced by attention and MLP respectively, and scalars $\alpha_{j}^{h,l,t}$, $\beta_{f}^{l,t}$ represent their corresponding coefficients computed through non-linear activations. $\text{LN}^t$ represents layernorm with scale and shift factors computed according to the state of the residual stream at timestep $t$. $W_e$, $W_u$ represent the embed/unembed linear projections, respectively, and $s_t$ represents the scale factor applied to the DiT output at timestep $t$ according to the diffusion noise schedule.

\section{Kernel Optimizations}
\label{appx-kernel-opt}

In this appendix, we provide additional details on Chipmunk's kernel optimizations. We structure this discussion into three primary categories: (1) efficiently identifying dynamic sparsity patterns in attention kernels, (2) optimizing column-sparse computations in MLP layers, and (3) reducing GPU memory overhead through effective caching strategies.

\subsection{Architecture Agnostic}
Chipmunk works on all modern GPUs with tensor cores. Although our CUDA kernels are optimized specifically for NVIDIA H100 GPUs, Chipmunk can be efficiently implemented on earlier or later GPU architectures by using architecture-agnostic kernels written in Triton. These kernels can also be tuned for optimal performance on other hardware. We include several example Triton kernels in our codebase for reference.

\subsection{Efficient Dynamic Sparsity Pattern Identification in Attention}
\label{appx-colsum}
We first address the challenge of efficiently identifying sparsity patterns in attention computations. As discussed previously (Section \ref{hw-efficiency}), Chipmunk employs fused kernels that simultaneously compute attention outputs and sparsity patterns via column sums (Alg. \ref{alg:colsumattention}). A direct summation of the unnormalized logits ($qk^T$) across rows is infeasible because row magnitudes can vary significantly without softmax normalization. Naively computing column sums directly on the post-softmax matrix is also impractical due to FlashAttention’s \cite{dao2022flashattentionfastmemoryefficientexact} incremental softmax, which never fully materializes intermediate softmax results within the kernel.

To overcome these limitations without resorting to slower, unfused kernels, we employ an approximation leveraging the slow-changing nature of activations across inference steps. Specifically, we reuse softmax normalization constants from the previous inference step to approximate column sums. Although slightly stale, these constants remain effective due to the incremental changes between steps. Thus, our fused kernel outputs both the correctly normalized attention result and an approximate column sum (normalized using previous constants) suitable for subsequent top-k sparsity selection.

\subsection{Optimizing Column-Sparse Delta MLP Computations}
\label{appx-mlp-opt}
Next, we detail kernel optimizations specific to column-sparse delta computations within the MLP layers. Here, the static nature of MLP value vectors (rows of weight matrix $W_2$) allows for additional optimization compared to attention, which has dynamic vectors projected from token representations.

\subsubsection{Computing MLP Step-Deltas Efficiently in One Pass}
Unlike attention, MLP step-deltas can be efficiently computed in a single step rather than requiring a subtraction and subsequent addition of vectors (Alg. \ref{alg:colsparsedeltamlp}). Given cached activations and MLP outputs, we directly compute sparse cross-step deltas as follows:

\begin{enumerate}
    \item Compute sparse intermediate activations.
    \item Compute the difference against the cached activation.
    \item Multiply this sparse delta by the static value vectors (rows of $W_2$).
    \item Directly accumulate this result into the cached output.
\end{enumerate}

This reduces computational overhead compared to the two-step subtraction-addition method required for dynamic attention vectors, but introduces additional challenges in kernel optimization.

\subsubsection{Persistent Grid and Warp-Specialization for Complex Epilogues}
The first GEMM kernel in Chipmunk's MLP delta computation involves a complex epilogue due to the combined computational steps of delta computation and memory operations (Alg. \ref{alg:colsparsedeltamlp}). To optimize these, we find the combination of persistent grids and warp-specialization to be particularly effective:

\begin{enumerate}
    \item \textbf{Persistent Grid Kernels:} One threadblock is launched per GPU Streaming Multiprocessor (SM), allowing each threadblock to iterate over multiple work tiles.
    \item \textbf{Warp-Specialization:} Within each threadblock, separate warp groups are assigned to compute/data loading operations, allowing better overlap between computation and memory operations.
\end{enumerate}

This combination allows the overlap of the producer warpgroup’s memory loading prologue with the consumer warpgroups' high latency epilogue operations.

\subsubsection{Custom Kernel for Efficient Cache Writeback}

We also found the most time-consuming step in the first MLP GEMM epilogue to be the scattering of activation cache updates into global memory. To address this, we fuse this memory-bound cache writeback operation into the second GEMM operation, which is compute-bound. Specifically, we utilize the CUDA driver API to allocate streaming multiprocessors (SMs) to a custom kernel implementing the cache writeback operation, while using the rest of the SMs for the GEMM. We compute the number of SMs to allocate based on the degree of wave quantization so that this does not impact the runtime of the GEMM—it just repurposes any leftover SMs. Our custom cache writeback kernel uses the TMA-based reduction PTX instructions (\texttt{cp.reduce.async.bulk}) to perform large atomic updates into global tensors.

\subsection{Minimizing GPU Memory Overhead from Activation Caches}
Finally, Chipmunk stores activation caches MLP and attention layers, making memory efficiency critical—particularly in single-GPU workloads with large sequence lengths (e.g., HunyuanVideo with 118k tokens per video). Each attention layer requires caching (1) boolean masks indicating active [192, 1] columns, and (2) activation outputs from the previous inference step.

We implement two optimizations to reduce GPU memory footprint:

\begin{enumerate}
    \item \textbf{Bitpacked Sparsity Masks:} Standard boolean masks (torch.bool) consume one byte per entry. With a torch-compiled bitpacking function, we reduce memory usage by 8x, while incurring negligible computational overhead.
    \item \textbf{CPU Offloading with Double-Buffered Communication:} We preallocate pinned (page-locked) CPU tensors and implement double-buffering on the GPU. This approach reduces GPU memory and communication overhead by overlapping GPU computations of the current layer with simultaneous transfers of the next layer’s sparsity masks and activations from CPU to GPU memory.
\end{enumerate}

\begin{table}[h]
\caption{Memory usage comparison between naive and optimized implementations.}
\centering
\begin{tabular}{|l|r|r|r|}
\hline
\textbf{} & \textbf{Naive} & \textbf{Optimized} & \textbf{Memory Reduction} \\
\hline
Sparsity Mask Cache & 104 GB & 13 GB & 8x \\
Activation Cache & 43 GB & 2.8 GB & 15x \\
Column-Sum Intermediate State & 668 GB & 3.5 GB & 192x \\
\hline
\end{tabular}
\label{tab:memory-optimizations}
\end{table}

\section{Chipmunk Algorithm}

In this appendix, we describe Chipmunk's algorithms in detail. The algorithms described here are for reference-correct implementations, rather than optimized for speed. In practice, operations in these algorithms implemented in optimized CUDA kernels as described in Section \ref{method} and Appendix \ref{appx-kernel-opt}.

\begin{figure}[htbp]
\centering
\begin{minipage}{\textwidth}
\begin{lstlisting}[style=algo,caption={\textbf{Voxel Reordering.} Reorders tokens so that contiguous chunks correspond to coherent 3D voxels, improving the quality of subsequent column-sparse approximations.},label={alg:voxel}]
# x            : (b, t, h, w, d)     3D tokens  
# vt, vh, vw   : int, int, int       Voxel shape

def voxel_order(x, vt, vh, vw):
    return rearrange(
        x,
       'b (tc vt) (hc vh) (wc vw) d -> b tc hc wc vt vh vw d',
        vt=vt, vh=vh, vw=vw
    )

def reverse_voxel_order(x, vt, vh, vw):
    return rearrange(
        x,
       'b tc hc wc vt vh vw d -> b (tc vt) (hc vh) (wc vw) d',
        vt=vt, vh=vh, vw=vw
    )
\end{lstlisting}
\end{minipage}
\end{figure}

\begin{figure}[htbp]
\centering
\begin{minipage}{\textwidth}
\begin{lstlisting}[style=algo,floatplacement=p,caption={\textbf{Chipmunk Attention.} Computes attention outputs by interleaving dense and sparse-delta steps. Dense steps initialize sparsity patterns and caches, while sparse steps selectively recompute attention interactions based on cached indices and activations.},label={alg:chipmunkattention}]
def chipmunk_attn(q, k, v, is_dense_step):
    # q, k, v       : (b, h, n, e)
    # is_dense_step : bool
    # returns o     : (b, h, n, e)
    
    if is_dense_step:
      o, cs, m, l  = colsum_attn(
                       q, k, v,
                       prev_m, prev_l
                     )
      inds, counts = topk(cs, dim=-1)
      prev_m       = m
      prev_l       = l
      o_cache      = colsparse_delta_attn(
                         q, k, v,
                         inds, counts,
                         o, o_scale = -1
                       )

    else:
      o            = colsparse_delta_attn(
                         q, k, v,
                         inds, counts,
                         o_cache, o_scale = 1
                       )
    
    return o
\end{lstlisting}
\end{minipage}
\end{figure}

\begin{figure}[htbp]
\centering
\begin{minipage}{\textwidth}
\begin{lstlisting}[style=algo,caption={\textbf{Column Sparse Delta Attention.} Efficiently recomputes sparse attention interactions defined by previously cached indices.},,label={alg:colsparsedeltaattn}]
def colsparse_delta_attn(q, k, v, inds, counts, o_cache, o_scale):
    # q, k, v       : (b, h, n, e)
    # inds          : (b, h, n/c, topk)   sparse indices per Q-tile
    # counts        : (b, h, n/c)         number active per Q-tile
    # o_cache       : (b, h, n, e)
    
    # c             : q  tile size
    # ck            : kv tile size
    
    o = o_cache.clone()
    for i in range(0, n, c):
        q_tile        = q[:, :, i : i+c]           # (b, h, c, e)
        o_tile        = o[:, :, i : i+c]           # (b, h, c, e)
        idx_q         = inds[:, :, i//c, :]        # (b, h, topk)
        counts_amt    = counts[:, :, i//c]         # (b, h, 1)
    
        for j in range(0, counts_amt, ck):
            idx_tile  = idx_q[..., j : j+ck]       # (b, h, ck)
            k_tile    = k.gather(idx_tile, dim=-2) # (b, h, ck, e)
            v_tile    = v.gather(idx_tile, dim=-2) # (b, h, ck, e)
            qk        = q_tile @ k_tile.T(-2, -1)  # (b, h, c, ck)
            p         = online_softmax(qk)         # (b, h, c, qk)
            o_tile   += o_scale * (p @ v_tile)     # (b, h, c, e)
    
    return o
\end{lstlisting}
\end{minipage}
\end{figure}

\begin{figure}[htbp]
\centering
\begin{minipage}{\textwidth}
\begin{lstlisting}[style=algo,caption={\textbf{Column Sum Attention.} Computes standard attention along with column-wise sums of attention probabilities, which are used for dynamically identifying sparsity patterns.},label={alg:colsumattention}]
def colsum_attn(q, k, v, prev_m, prev_l):
    # q, k, v   : (b, h, n, e)
    # prev_m    : (b, h, n, 1)
    # prev_l    : (b, h, n, 1)
    # returns o : (b, h, n, e)
    #        cs : (b, h, n/c, n)  column-chunk sums
    
    qk       = q @ k.T(-2, -1) / sqrt(e)         # (b, h, n, n)
    p        = softmax(qk, dim=-1)               # (b, h, n, n)
    o        = p @ v                             # (b, h, n, e)
    
    p_approx = exp(qk - prev_m) / prev_l         # (b, h, n, n)
    cs       = rearrange(                        # (b, h, n/c, n)
                 p_approx, 
                'b h (nc c) m -> b h nc c m',
                 c=c
               ).sum(dim=-2)
    
    return o, cs
\end{lstlisting}
\end{minipage}
\end{figure}

\begin{figure}[htbp]
\centering
\begin{minipage}{\textwidth}
\begin{lstlisting}[style=algo,caption={\textbf{Chipmunk MLP.} Performs MLP layer computations by alternating dense and sparse-delta steps. Dense steps compute full activations and initialize caches, and sparse steps efficiently update only the activations that exhibit significant changes across inference steps.},label={alg:chipmunkmlp}]
def chipmunk_mlp(x, w1, w2, is_dense_step):
    # x             : (b, n, d)
    # w1, w2        : (f, d)
    # is_dense_step : bool
    # returns o     : (b, n, d)
    
    if is_dense_step:
      o, preact, act = mlp(x, w1, w2)
      tm_cache       = rearrange(
                         preact,
                        'b (nc c) d -> b nc c d',
                         c=192
                       ).mean(dim=-2)
      act_cache      = act.clone()
      mlp_cache      = o.clone()

    else:
      inds, counts   = compute_mlp_indices(
                         x,
                         w1,
                         tm_cache
                       )
      o              = colsparse_delta_mlp(
                         x,
                         w1, w2,
                         self.inds, self.counts,
                         act_cache, o_cache
                       )
    
    return o
\end{lstlisting}
\end{minipage}
\end{figure}

\begin{figure}[htbp]
\centering
\begin{minipage}{\textwidth}
\begin{lstlisting}[style=algo,caption={\textbf{Column Sparse Delta MLP.} Computes sparse deltas for MLP outputs by selectively recomputing only significantly changed neuron activations.},label={alg:colsparsedeltamlp}]
def colsparse_mlp_delta(x, w1, w2, inds, counts, a_cache, o_cache):
    # x, w1, w2   : (b, n, d), (f, d), (f, d)
    # inds        : (b, n/c, topk)    active neurons per token-tile
    # counts      : (b, n/c)          number active per token-tile
    # a_cache     : (b, f, n)         cached activations
    # o_cache     : (b, n, d)         cached MLP output
    #
    # c           : M tile size
    # cn          : N tile size
    # ck          : K tile size
    
    # ---- GEMM1 : delta = x @ W1[idx, :].T - a_cache  ----
    tasks = [(i, j) for i in range(0,         n, c)
                    for j in range(0, counts[i], cn)]
    for (i, j) tasks:
        idx_tile     = inds[:, i//c, j : j+cn]         # (b, cn)
        a_tile       = a_cache[idx_tile].T             # (b, c, cn)
        pre          = zeros_like(a_tile)              # (b, c, cn)
    
        for k in range(0, d, ck):
            x_tile   = x[:, i: i + c, k: k + ck]       # (b, c, ck)
            w1_tile  = w1[idx_tile, k : k + ck]        # (b, cn, ck)
            pre     += x_tile @ w1_tile.T              # (b, c, cn)
    
        a            = act_fn(pre + w1.bias[idx_tile]) # (b, c, cn)
        delta        = a - a_cache                     # (b, c, cn)
        a_cache.scatter_(-2, idx, a.T)                 # (b, cn, c)
    
    # ---- GEMM2 : o_cache += delta @ W2[idx, :]  ----
    tasks = [(i, j) for i in range(0, n, c)
                    for j in range(0, d, cn)]
    for (i, j) tasks:
        o_tile       = o_cache[:, i: i + c]            # (b, c, cn)
    
        for k in range(0, counts[i], ck):
            idx_tile = inds[:, i//c, k: k + ck]        # (b, ck)
            d_tile   = delta[:, i: i + c, k: k + ck]   # (b, c, ck)
            w2_tile  = w2[idx_tile, j : j + cn]        # (b, ck, cn)
            o_tile  += delta @ w2_tile                 # (b, c, cn)
    
    return o_cache                                     # (b, n, d)
\end{lstlisting}
\end{minipage}
\end{figure}

\begin{figure}[htbp]
\centering
\begin{minipage}{\textwidth}
\begin{lstlisting}[style=algo,caption={\textbf{Computing MLP Sparse Indices.} Identifies neurons in MLP layers with the largest cross-step activation changes.},label={alg:computemlpindices}]
def compute_mlp_indices(x, w1, tm_cache):
    # x                : (b, n, d)
    # w1               : (f, d)
    # tm_cache         : (b, n/c, f)
    #
    # returns inds   : (b, n/c, topk)  top neurons per token-group
    #         counts : (b, n/c, 1)     active neurons per token-group
    
    tm           = rearrange(                       # (b, n/c, d)
                     x,
                    'b (nc c) d -> b nc c d',
                     c=c
                   ).mean(dim=-2)                   
    tm           = tm @ w1.T                        # (b, n/c, f)
    delta        = (tm - tm_cache).abs()            # (b, n/c, f)
    inds, counts = delta.topk(dim=-1).indices       # (b, n/c, topk)
    
    return inds, counts
\end{lstlisting}
\end{minipage}
\end{figure}
\FloatBarrier

\section{Evaluations}

\subsection{Hyperparameters}\label{appx-hyperparameters}
In this section, we expand upon the hyperparameters of all methods used to create the tables.

\begin{itemize}
    \item \textbf{TeaCache: } The threshold parameter is set to 0.78 for FLUX.1-dev, 0.2 for WAN2.1, and 0.65 for HunyuanVideo. We implement TeaCache by copying the \href{https://github.com/ali-vilab/TeaCache/}{reference repository} implementation.
    \item \textbf{Sliding Tile Attention: }We approximated the tile size to approximately cover 58\% sparsity, as described in their paper \cite{zhang2025fastvideogenerationsliding}. For speedup numbers, we run \href{https://github.com/hao-ai-lab/FastVideo/blob/main/csrc/sliding_tile_attention/st_attn/st_attn_h100.cu}{their kernel} at this sparsity level and measure the kernel duration, replacing the existing attention runtime with the new number.
    \item \textbf{DiTFastAttn: }The default hyperparameters available at the \href{https://github.com/xdit-project/xDiT}{xDiT GitHub repository} were used: window size=512, number of calibrations=4, and with caching enabled.
    \item \textbf{ToCa: }The default hyperparameters suggested in their paper were used (N=2, R=90\%) \cite{zou2025acceleratingdiffusiontransformerstokenwise}.
    \item \textbf{Chipmunk: }We apply 84\% attention sparsity and 70\% MLP sparsity for FLUX.1-dev. For text-to-video generation we use 95\% attention sparsity on HunyuanVideo and 82\% attention sparsity on WAN2.1. Since MLP runtime accounts for a very small percentage of wall clock time in both HunyuanVideo and WAN2.1, we only apply sparse attention deltas to achieve the best speed-quality tradeoff. For WAN2.1, when we stack Chipmunk with STA, we use a configuration of STA with a sliding tile size of 3 voxels, where each voxel is a (4,6,8) chunk of tokens. This corresponds to roughly 7\% of the total attention mask.
    \item \textbf{Step Caching: } For entries marked Chipmunk + Step Caching, we use a simple uniform schedule we found to approximate the behavior of a number of adaptive scheduling methods. In the middle $W$ steps of the diffusion process, we only compute every $n$th step, skipping all others by reusing the last computed model output. We use $W=30$ and $n=4$, which corresponds to a roughly 1.8x speedup with 50 total steps. Thus, each step is either (i) fully skipped, (ii) partially sparse (Chipmunk), or (iii) fully dense.

\end{itemize}

\subsection{HunyuanVideo}\label{appx-evaluations}
\begin{figure}
    \centering
    \includegraphics[width=1\linewidth]{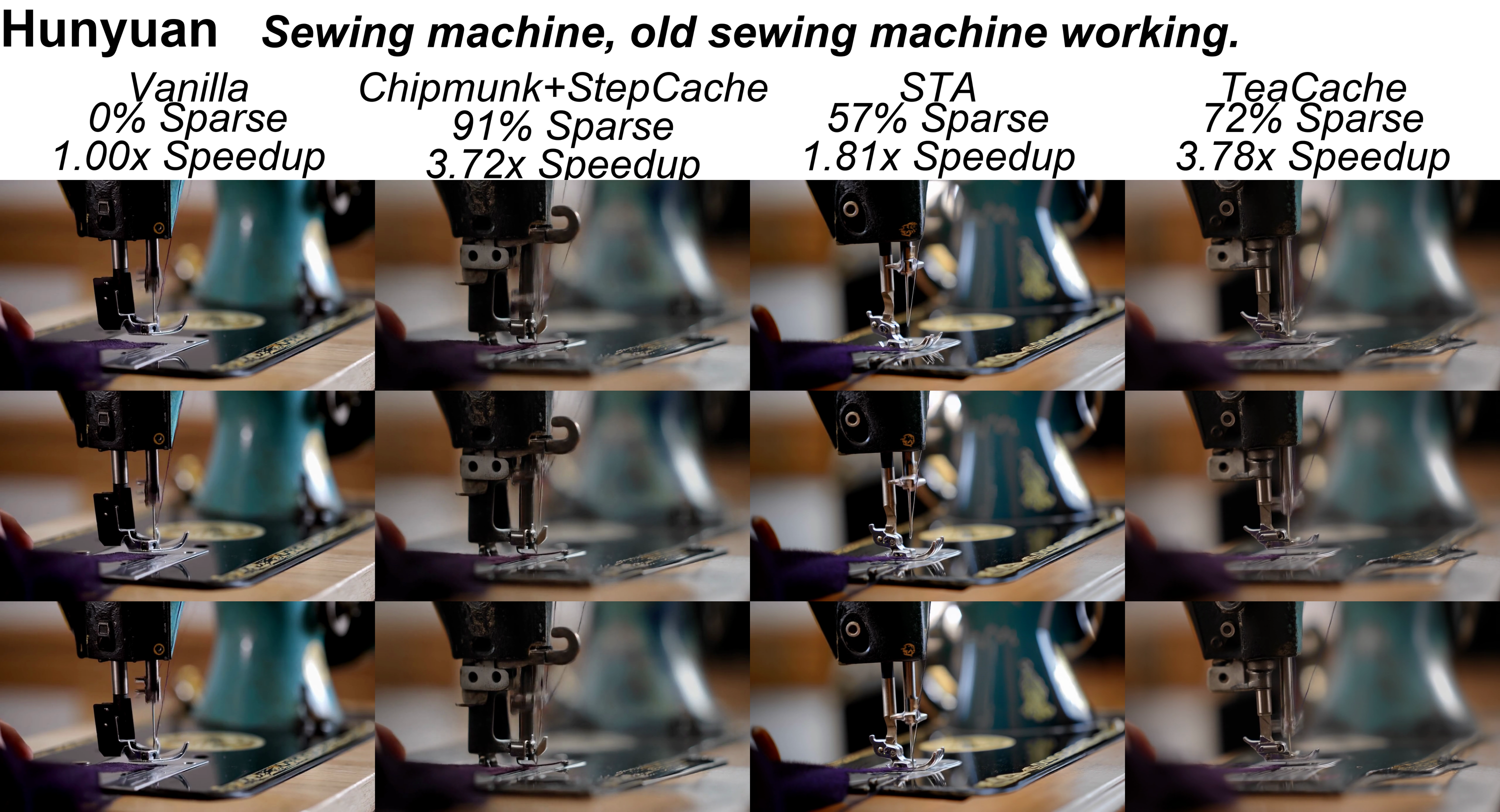}
    \caption{Qualitative video generation comparison across different methods for HunyuanVideo}
    \label{fig:large-qualitative-grid-hunyuan}
\end{figure}

We include additional qualitative studies on text-to-video generation quality in Figure \ref{fig:large-qualitative-grid-hunyuan}.

\subsection{WAN2.1}
\begin{figure}
    \centering
    \includegraphics[width=\linewidth]{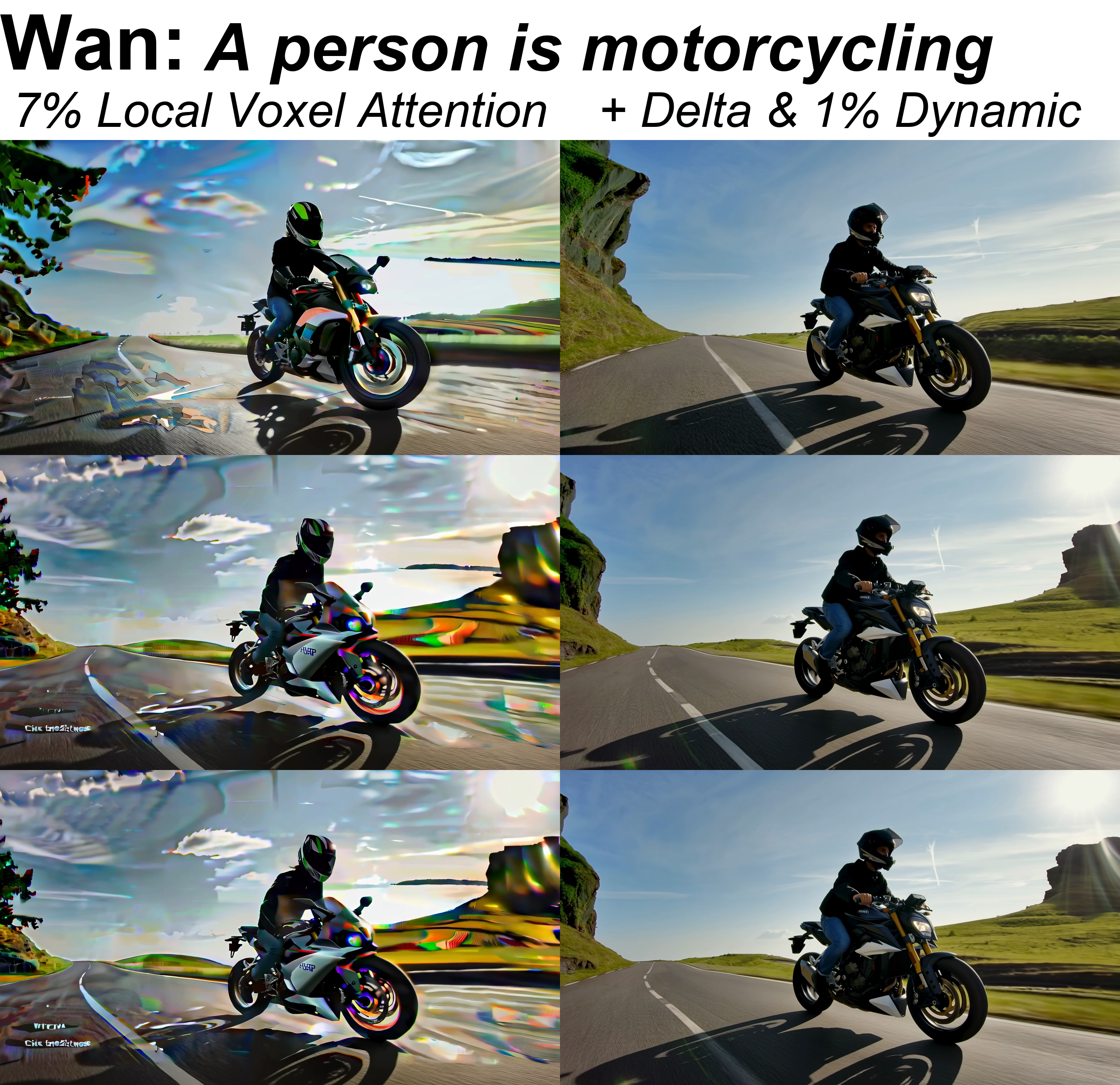}
    \caption{Ablation of adding deltas (to reuse attention interactions that are not recomputed) and 1\% dynamic TopK attention mask on top of static local voxel attention. We find that at high sparsity levels, using only a static local mask results in artifacts and object warping. Adding deltas and just 1\% of the dynamically selected top attention interactions significantly improves quality. \textit{Note: Because the 3D dimensions do not divide evenly into 3D voxels, both configurations shown above also use full attention to and from the remainder slice in each dimension.}}
    \label{fig:wan-local-plus-delta-dynamic}
\end{figure}

\begin{figure}
    \centering
    \includegraphics[width=\linewidth]{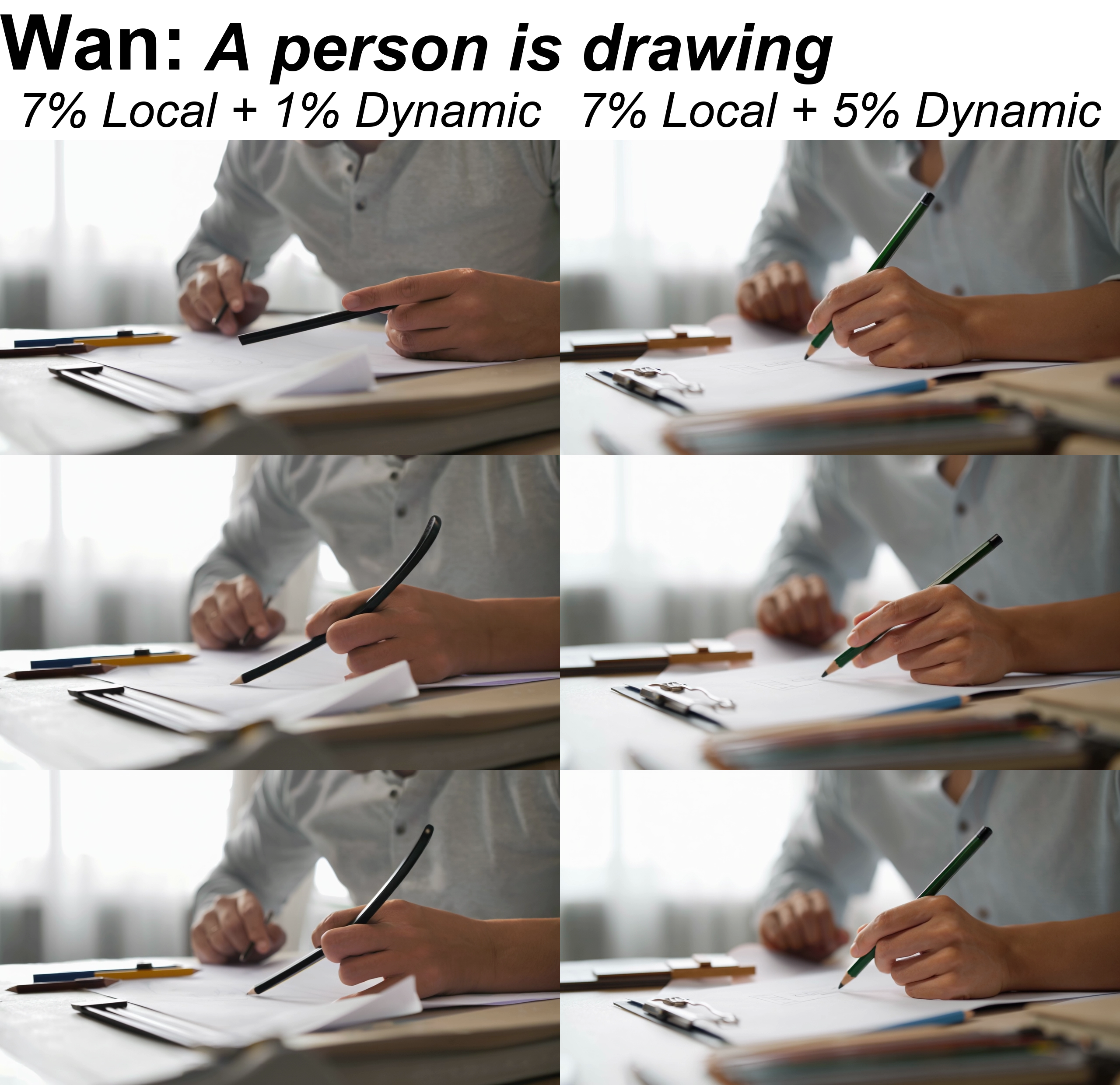}
    \caption{Comparing the quality of WAN2.1 generation at 1\% dynamic top-k attention interactions and 5\% dynamic top-k attention interactions. On the left, we see that at high sparsity levels, detailed objects such as the pencil and hands begin to experience warping. We then see on the right that increasing the number of dynamically selected top-k attention interactions from 1\% to 5\% restores high quality generation of the pencil and hands.}

    \label{fig:wan-top-keys-comparison}
\end{figure}

\begin{figure}
    \centering
    \includegraphics[width=1\linewidth]{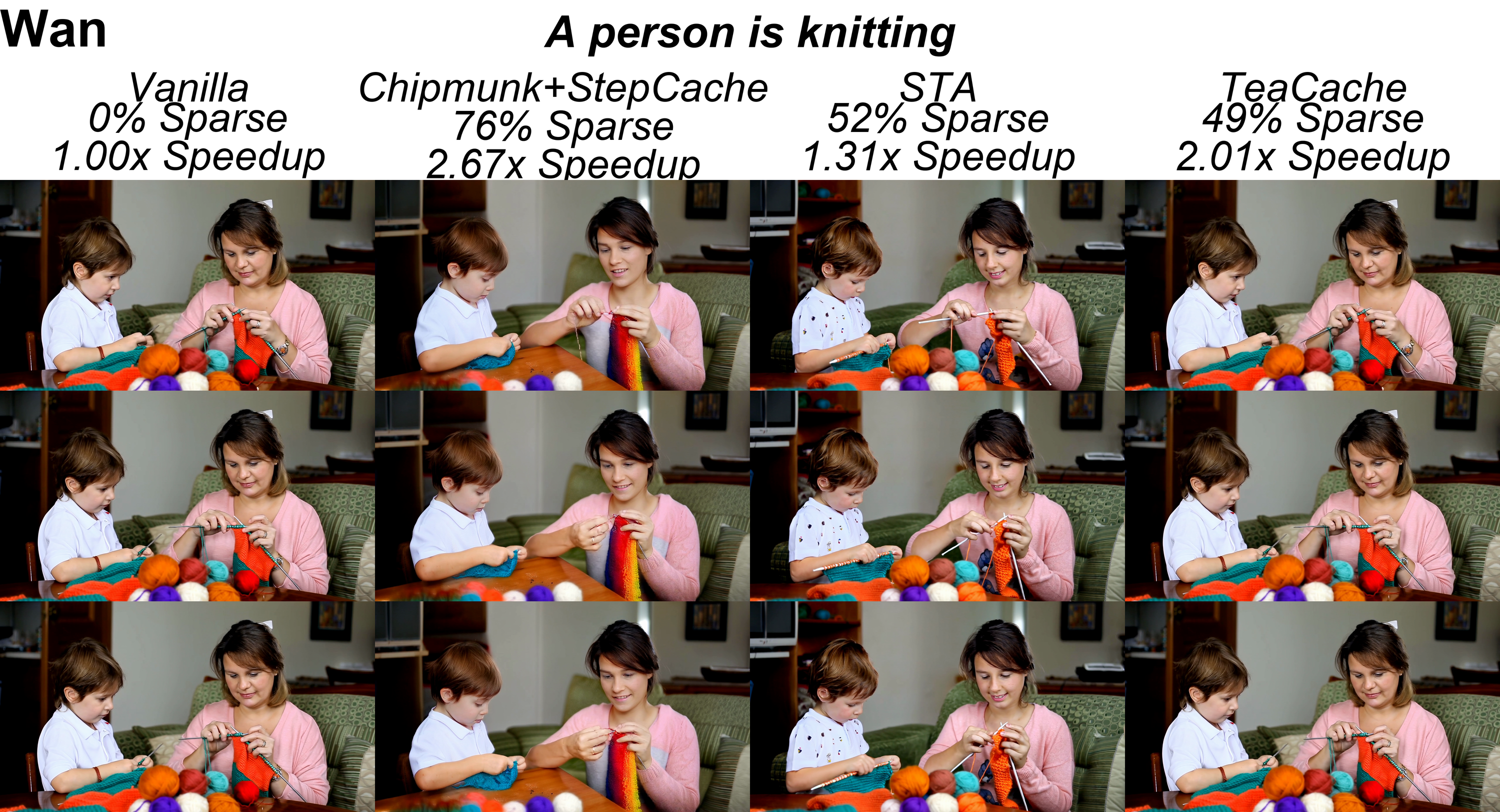}
    \includegraphics[width=1\linewidth]{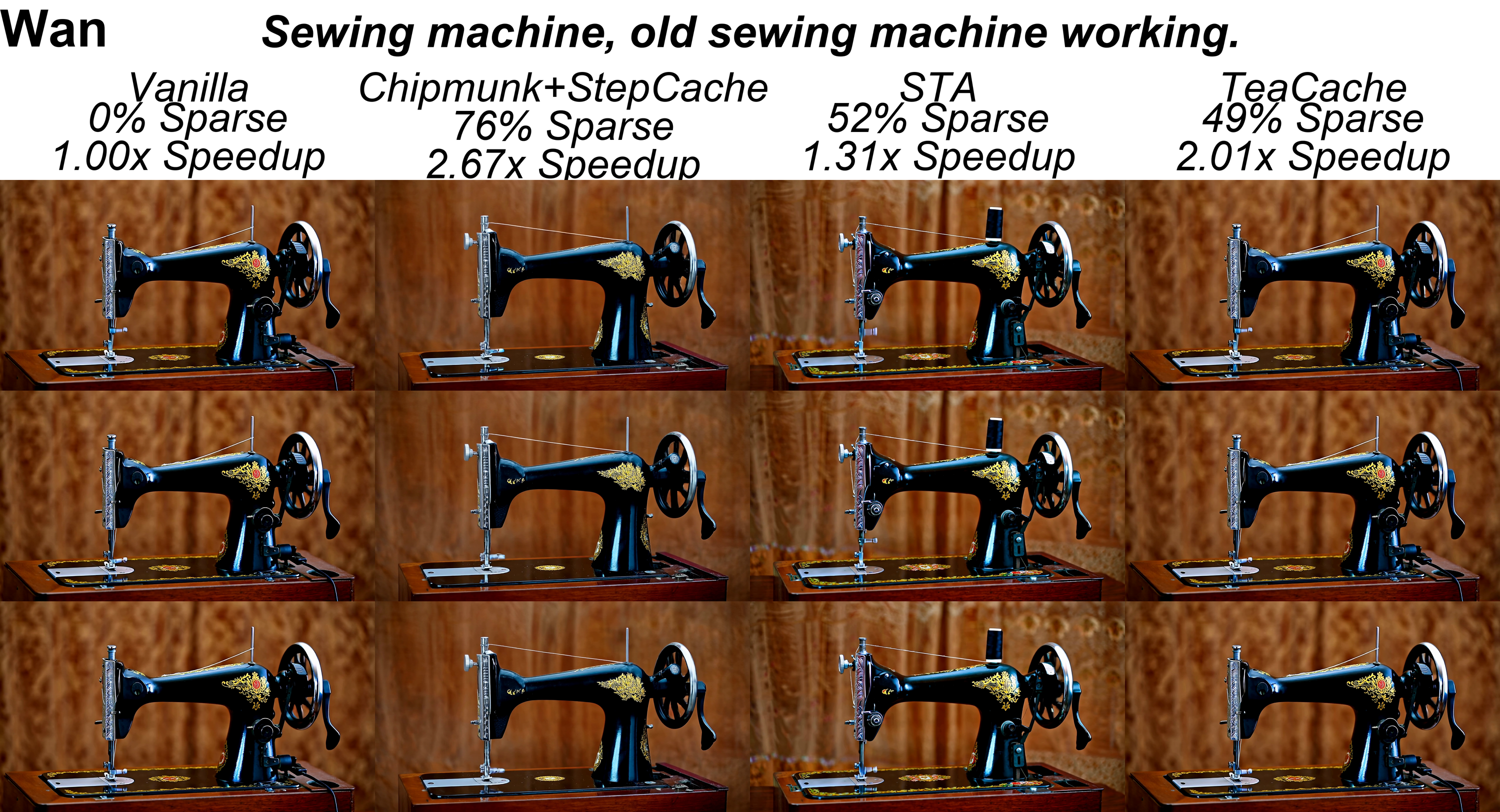}
    \caption{Qualitative video generation comparison across different methods for WAN2.1.}
    \label{fig:large-qualitative-grid-wan}
\end{figure}

We include additional qualitative studies on text-to-video generation quality in Figures \ref{fig:wan-local-plus-delta-dynamic}, \ref{fig:wan-top-keys-comparison}, and \ref{fig:large-qualitative-grid-wan}. In Fig. \ref{fig:wan-local-plus-delta-dynamic}, we test the effect of reusing non-recomputed attention interactions from a previous step and using dynamically identified sparsity patterns. On the left, we observe that only using a static local mask at high sparsity levels introduces quality degradation and object warping. On the right, we find that reusing attention interactions and adding only 1\% of dynamically selected top attention interactions restores significant quality. We also study the speed-quality tradeoff at high levels of sparsity in Fig. \ref{fig:wan-top-keys-comparison}. At high levels of sparsity, such as 92\% on the left, we observe warping on detailed objects such as a hand making fine motor movements to draw on paper. We find that slightly increasing the number of dynamically selected attention interactions to be recomputed, from 1\% to 5\%, significantly improves quality.

\subsection{FLUX.1-dev}

\begin{figure}
    \centering
    \includegraphics[width=\linewidth]{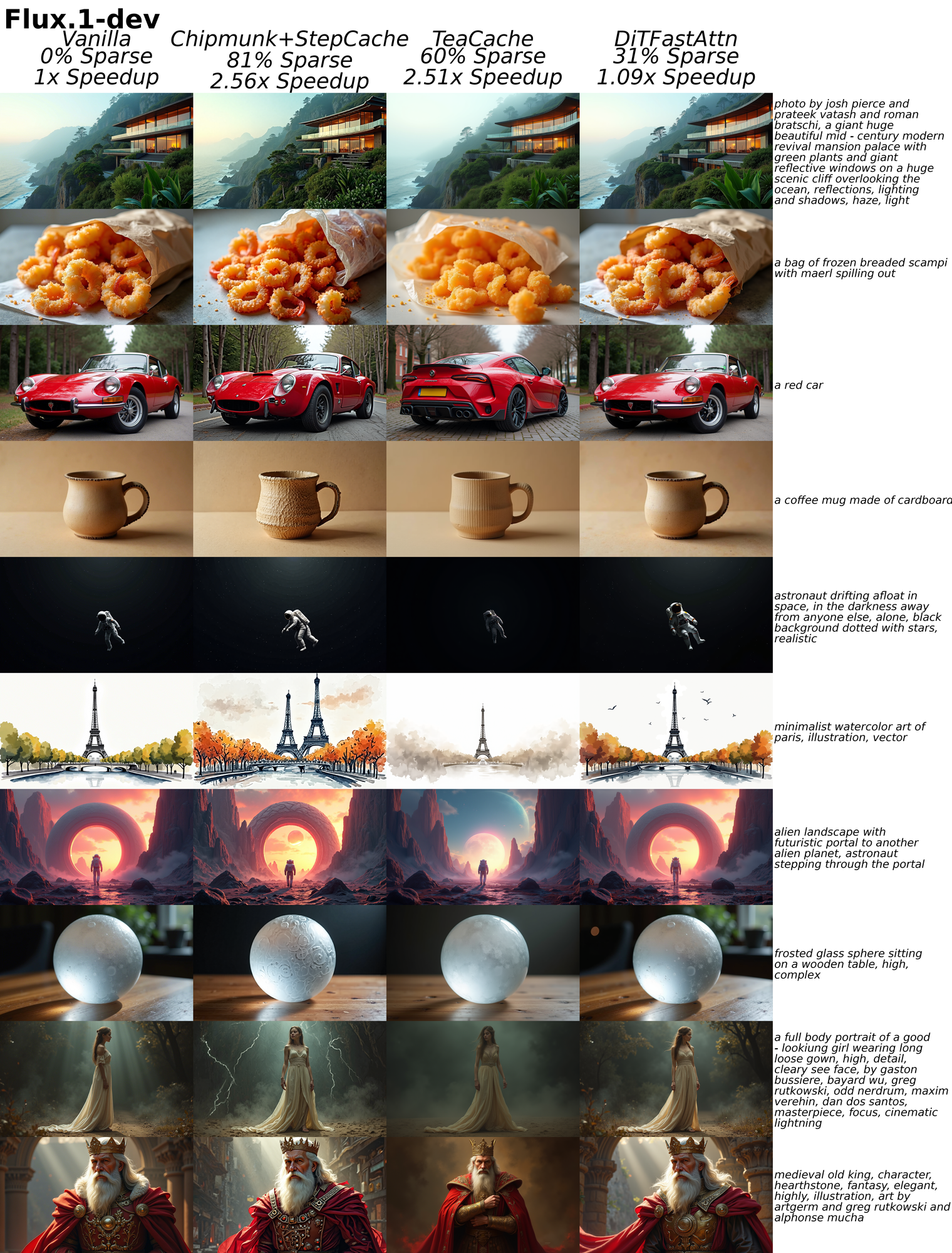}
    \caption{Images on 1280x768 FLUX.1-dev evaluated on different captions and prompts randomly sampled from the ImageReward dataset. On the left, we have vanilla reference images. Naively skipping steps (third column) introduces significant artifacts, such as a blurry images and a loss of detail. Chipmunk preserves aesthetic quality but may change some details of the image (despite maintaining prompt adherence). DiTFastAttn also achieves high quality but only a minor speedup.}

    \label{fig:flux_methods_and_prompts}
\end{figure}

\paragraph{Evaluation across prompts.} Using the methods of the main results table in the paper body, we generate images across a variety of prompts and methods (Fig. \ref{fig:flux_methods_and_prompts}). We observe that naively skipping steps may impair visual quality due to blur and loss of detail. We find Chipmunk preserves aesthetic quality but may change some details of the image while maintaining strong adherence to prompts. Although DiTFastAttn has strong quality, it only achieves a minor speedup.

\begin{figure}
    \centering
    \includegraphics[width=\linewidth]{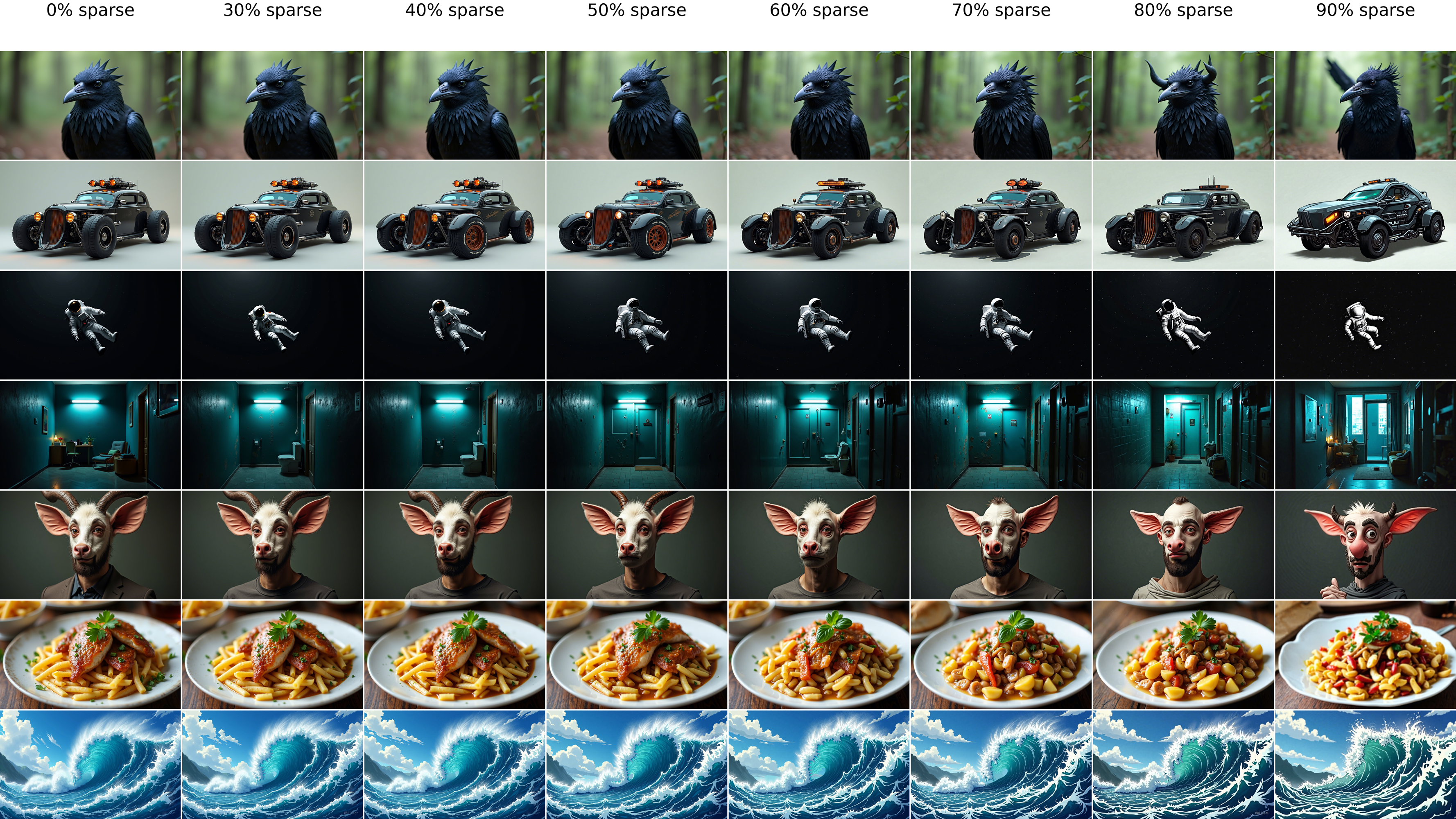}
    \caption{Speed-quality tradeoff on 1280x768 images generated on FLUX.1-dev with 50 steps at varying levels of Chipmunk sparsity. The images maintain strong quality at high levels of sparsity reaching 70-80\%. Beyond 80\%, significant artifacts are introduced and visual quality noticeably degrades resulting in blurry images or loss of detail. A single value was used for both attention and MLP sparsity. Prompts are listed below:}
        \begin{itemize}
    \item "anthropomorphic crow werecreature, photograph captured in a forest"
    \item "a concept art of a vehicle, cyberpunk"
    \item "astronaut drifting afloat in space, in the darkness away from anyone else, alone, black background dotted with stars, realistic"
    \item "photo of a interior taken with a cheap digital camera at night flash lighting"
    \item "A realistic photo of a man with big ears"
    \item "delicious plate of food"
    \item "tumultuous plunging waves, anime, artwork, studio ghibli, stylized, in an anime format"
    \item "an alien planet viewed from space, extremely, beautiful, dynamic, creative, cinematic"
    \end{itemize}

    \label{fig:flux_speed_quality_tradeoff}
\end{figure}

\paragraph{Evaluation across speed-quality tradeoff.} We evaluate multiple values of MLP and attention sparsity, ranging from 0\% to 90\%, to better understand how the MLP \& attention sparsity parameters modulate the speed-quality tradeoff (Fig. \ref{fig:flux_speed_quality_tradeoff}). The images maintain strong quality levels of sparsity reaching 70-80\%. Beyond 80\%, significant artifacts are introduced.


\section{Extended Related Work}

\textbf{Towards Few-Step Diffusion Models.} Early diffusion models required hundreds to thousands of denoising steps per generated sample, originating from the foundational work of Sohl-Dickstein et al. \cite{sohldickstein2015deepunsupervisedlearningusing} and Ho et al. \cite{ho2020denoisingdiffusionprobabilisticmodels}. Subsequent methods significantly reduced inference steps by enhancing sampling efficiency: DDIM introduced a non-Markovian forward diffusion process that decouples training and sampling steps, while the DPM-Solver family achieved substantial speedups by utilizing dedicated diffusion ODE solvers without retraining \cite{song2022denoisingdiffusionimplicitmodels, lu2022dpmsolverfastodesolver}. Another line of research has specifically targeted training processes optimized for single-step inference. For instance, Rectified Flow \cite{liu2022flowstraightfastlearning} learns straight-line mappings from noise to data, and Consistency Models \cite{song2023consistencymodels} directly model single-step noise-to-data transformations from arbitrary points along the noising trajectory. Additionally, step-distillation techniques, such as Progressive Distillation \cite{salimans2022progressivedistillationfastsampling}, efficiently train few-step student models to approximate the longer sampling trajectories of multi-step teacher models. Complementing these approaches, which allocate computational resources uniformly once the number of steps is determined, Chipmunk studies an orthogonal dimension of efficiency: dynamically allocating computation within individual inference steps by selectively recomputing only the most important activations.

\textbf{Efficient Attention Approximations.} The quadratic complexity of self-attention mechanisms has driven extensive research into efficient approximations. Prior works have explored many strategies, including low-rank approximations \cite{wang2020linformerselfattentionlinearcomplexity}, random-feature projections  \cite{choromanski2022rethinkingattentionperformers}, locality-sensitive hashing or local attention masks \cite{kitaev2020reformerefficienttransformer, zaheer2021bigbirdtransformerslonger}, and combined sparse-plus-low-rank decompositions \cite{chen2021scatterbrainunifyingsparselowrank, arora2025simplelinearattentionlanguage}. Recent studies on video and diffusion models have adopted static sliding-window attention masks \cite{zhang2025fastvideogenerationsliding, yuan2024ditfastattnattentioncompressiondiffusion} and quantized attention computations \cite{zhang2025sageattentionaccurate8bitattention}. Chipmunk focuses on a dynamic sparse attention approximation to speedup DiTs and can be complemented with other techniques such as low-rank approximations or static mask patterns.

\textbf{Efficient MLP Approximations.} Conditional computation strategies such as Mixture-of-Experts \cite{shazeer2017outrageouslylargeneuralnetworks, fedus2022switchtransformersscalingtrillion} selectively activate a subset of expert MLP modules per token, reducing FLOPs proportionally to top-$k$ gating decisions. More fine-grained neuron-level sparsity methods, such as contextual sparsity \cite{liu2023dejavucontextualsparsity}, dynamically select only the most relevant neurons during autoregressive decoding, guided by a lightweight prediction model. Chipmunk complements these existing conditional computation techniques with the addition of activation reuse across inference steps. Instead of purely gating experts or neurons, Chipmunk leverages multi-step inference to selectively recompute only the neurons exhibiting significant changes, while reusing the activations of non-activated neurons from a cache. While initially demonstrated for diffusion models, the general concept of caching and sparse recomputation has potential applicability in other multi-step inference contexts, such as autoregressive decoding.


\end{document}